%% file: main_arxiv.tex
\documentclass{article} 
\usepackage{arxiv,times}

\input{math_commands.tex}

\usepackage[utf8]{inputenc} 
\usepackage[T1]{fontenc}    
\usepackage{booktabs}       
\usepackage{amsfonts}       
\usepackage{nicefrac}       
\usepackage{microtype}      
\usepackage{doi}

\usepackage{hyperref}
\usepackage{amsmath}
\usepackage{url}
\usepackage{clipboard}
\usepackage{amsthm, bm}
\usepackage{wrapfig,lipsum}
\usepackage{verbatim}
\usepackage{mathtools}
\usepackage{tabularx}
\usepackage{siunitx}
\usepackage{dblfloatfix}

\newtheorem{theorem}{Theorem}[section]

\newtheorem{lemma}[theorem]{Lemma}

\newtheorem{definition}{Definition}

\newcommand{\xhdr}[1]{{\noindent\bfseries #1}.}

\def\eigvec{{\bm{\phi}}}
\def\eigvecnorm{\hat{{\bm{\phi}}}}

\title{Directional Graph Networks}
\renewcommand{\undertitle}{
Anisotropic aggregation in graph neural networks via directional vector fields
}

\author{Dominique Beaini\thanks{equal contribution}\\
Valence Discovery\\
Montreal, QC, Canada\\
\texttt{dominique@valencediscovery.com} \\
\And
Saro Passaro$^*$\\
 University of Cambridge \\
Cambridge, United Kingdom \\
\texttt{sp976@cam.ac.uk} \\
\And
Vincent Létourneau\\
Valence Discovery\\
Montreal, QC, Canada\\
\AND
William L. Hamilton \\
McGill University, MILA \\
Montreal, QC, Canada \\
\And
Gabriele Corso \\
University of Cambridge \\
Cambridge, United Kingdom \\
\And
Pietro Li\`{o} \\
 University of Cambridge \\
Cambridge, United Kingdom \\
}

\begin{document}

\maketitle

\begin{abstract}

The lack of anisotropic kernels in graph neural networks (GNNs) strongly limits their expressiveness, contributing to well-known issues such as over-smoothing. To overcome this limitation, we propose the first globally consistent anisotropic kernels for GNNs, allowing for graph convolutions that are defined according to topologicaly-derived directional flows.
First, by defining a vector field in the graph, we develop a method of applying directional derivatives and smoothing by projecting node-specific messages into the field. 
Then, we propose the use of the Laplacian eigenvectors as such vector field.
We show that the method generalizes CNNs on an $n$-dimensional grid and is provably more discriminative than standard GNNs regarding the Weisfeiler-Lehman 1-WL test.
We evaluate our method on different standard benchmarks and see a relative error reduction of 8\% on the CIFAR10 graph dataset and 11\% to 32\% on the molecular ZINC dataset, and a relative increase in precision of 1.6\% on the MolPCBA dataset. 
An important outcome of this work is that it enables graph networks to embed directions in an unsupervised way, thus allowing a better representation of the anisotropic features in different physical or biological problems. 

\end{abstract}

\begin{figure*}[ht]
\centering
 \includegraphics[width=\textwidth]{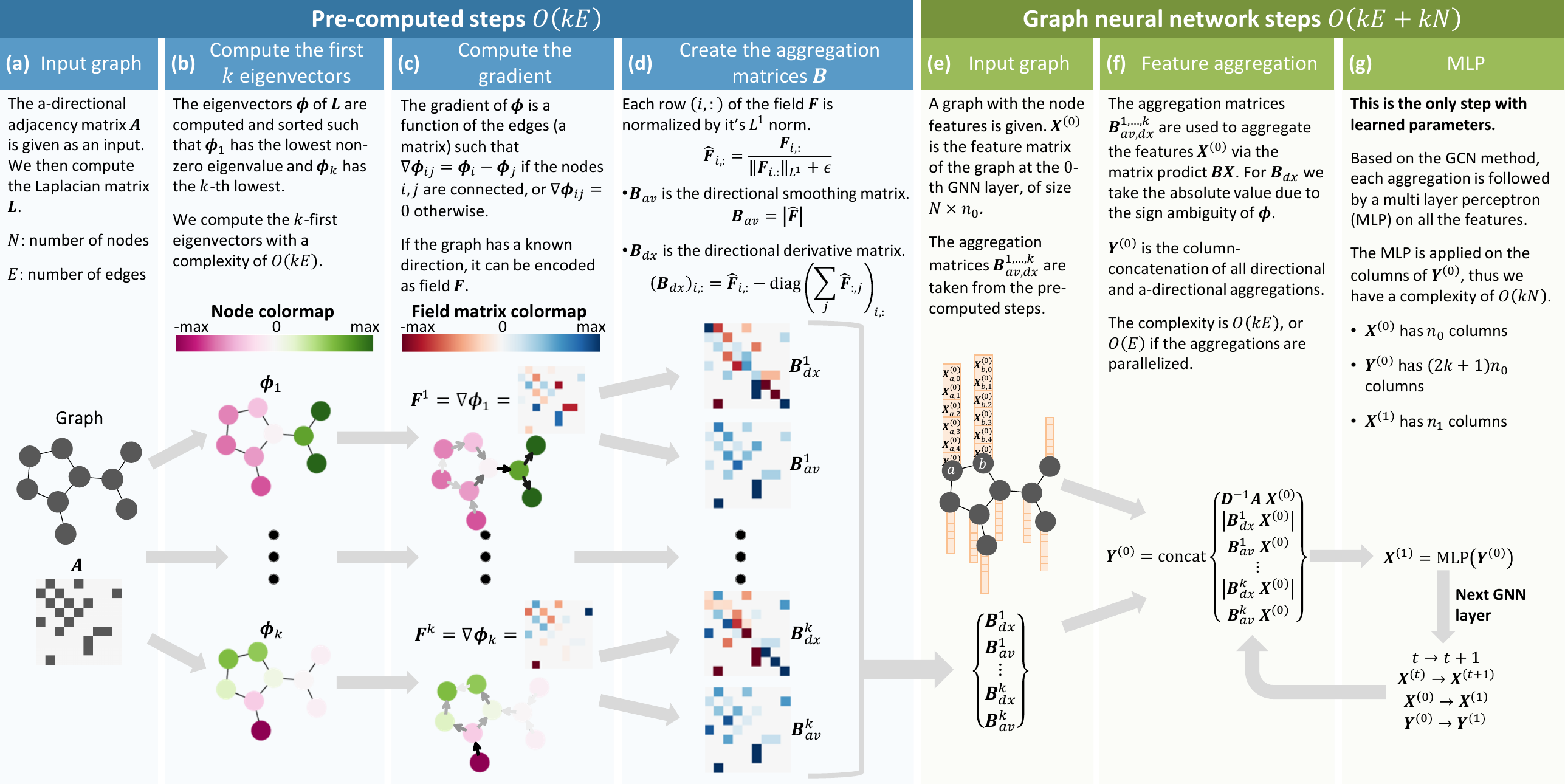}
 \vspace{-20pt}
 \caption{Overview of the steps required to aggregate messages in the direction of the eigenvectors.} 
\label{fig:full-method}
\end{figure*}

\section{Introduction}

One of the most important distinctions between convolutional neural networks (CNNs) and graph neural networks (GNNs) is that CNNs allow for any convolutional kernel, while most GNN methods are limited to symmetric kernels (also called isotropic kernels) \cite{kipf2016gcn,gilmer2017mpnn}. There are some implementations of asymmetric kernels using gated mechanisms \cite{bresson2017gatedGCN, velikovic2017gat}, motif attention \cite{peng_graph_2019_motif}, edge features \cite{gilmer2017mpnn}, port numbering \cite{sato2019approximation} or the 3D structure of molecules \cite{klicpera_directional_2019_dimenet}.

However, to the best of our knowledge, there are currently no methods that allow asymmetric graph kernels that are dependent on the full graph structure or directional flows. They either depend on local structures or local features. This is in opposition to images, which exhibit canonical directions: the horizontal and vertical axes. The absence of an analogous concept in graphs makes it difficult to define directional message passing and to produce an analogue of the directional frequency filters (or Gabor filters) widely present in image processing \cite{olah_overview_2020}. In fact, there is numerous evidence that directional filtering is fundamental image processing \cite{kang_deep_2017, antoine_two-dimensional_1996, yue_lu_finer_2005}.

We propose a novel idea for GNNs: use vector fields in the graph to define directions for the propagation of information. An overview of this framework is presented in figure \ref{fig:full-method}. Using this approach, the usual message-passing structure of a GNN is projected onto globally-defined directions so that the contribution of each neighbouring node $n_v$ is weighted by its alignment with the vector fields at the receiving node $n_u$. This enables our method to propagate information via directional derivatives or smoothing of the features.

\begin{figure*}[ht]
\centering
 \includegraphics[height=5.7cm]{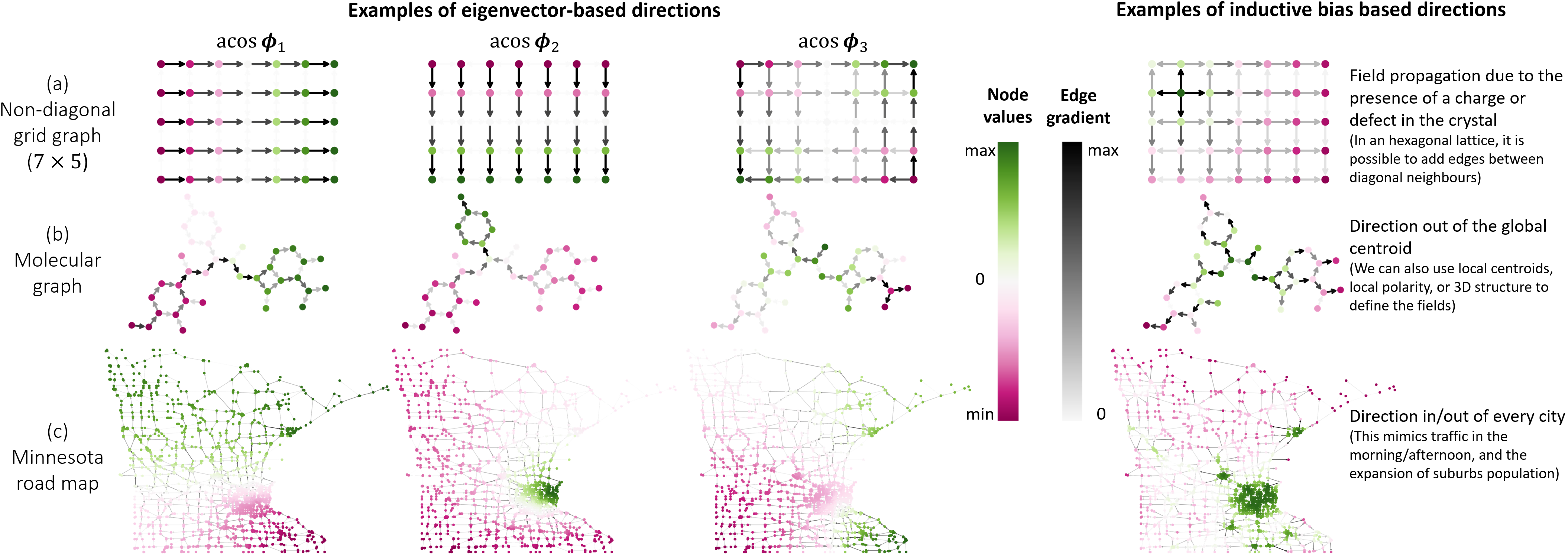}
 \vspace{-5pt}
 \caption{Possible directional flows in different types of graphs. The node coloring is a potential map and the edges represent the gradient of the potential with the arrows in the direction of the flow. The first 3 columns present the arcosine of the normalized eigenvectors ($\text{acos } \eigvecnorm$) as node coloring, and their gradients represented as edge intensity. The last column presents examples of inductive bias introduced in the choice of direction. (a) The eigenvectors 1 and 2 are the horizontal and vertical flows of the grid. (b) The eigenvectors 1 and 2 are the flow in the longest and second-longest directions. (c) The eigenvectors 1, 2 and 3 flow respectively in the South-North, suburbs to the city center and West-East directions. We ignore $\eigvec_0$ since it is constant and has no direction.} 
\label{fig:eig_vec_grad}
\end{figure*}

In order to define globally consistent directional fields over general graphs, we propose to use the gradients of the low-frequency eigenvectors $\eigvec_k$ of the graph Laplacian, since they are known to capture key information about the global structure of graphs \cite{chavel1984eigenvalues,chung1997spectral,Grebenkov_2013}. In particular, these eigenvectors can be used to define optimal partitions of the nodes in a graph, to give a natural ordering \cite{levy_laplace_beltrami_2006}, and to find the dominant directions of the graph diffusion process \cite{chung_discrete_2000,saerens2004principal}. Further, we show that they generalize the horizontal and vertical directional flows in a grid (see figure \ref{fig:eig_vec_grad}), allowing them to guide the aggregation and mimic the asymmetric and directional kernels present in computer vision. In fact, we demonstrate mathematically that our work generalizes CNNs, by reproducing all convolutional kernels of radius $R$ in an $n$-dimensional grid, while also bringing the powerful data augmentation capabilities of reflection, rotation or distortion of the directions. Additionally, we also prove that our directional graph networks (DGNs) are more discriminative than standard GNNs in regards to the Weisfeiler-Lehman 1-WL test, confirming an increase of expressiveness.

We further show that our DGN model theoretically and empirically allows for efficient message passing across distant communities, which counteracts the well-known problem of over-smoothing in GNNs. Alternative methods reduce the impact of over-smoothing by using skip connections \cite{luan2019break}, global pooling \cite{alon_bottleneck_2020}, or randomly dropping edges during training time \cite{rong_dropedge_2020}, but without solving the underlying problem. 

Our method distinguishes itself from other spectral GNNs since the literature usually uses the low frequencies to estimate local Fourier transforms in the graph \cite{levie_cayleynets_2018, xu_graph_2019}. Instead, we do not try to approximate the Fourier transform, but only to define a directional flow at each node and guide the aggregation.

We tested our method on 5 standard datasets from \cite{dwivedi2020benchmarking} and \cite{hu2020open}, using two types of architectures, and either using or ignoring edge features. In all cases, we observed state-of-the-art results from the proposed DGN, with relative improvements of 8\% on CIFAR10, 11-32\% on ZINC, 0.8\% on MolHIV and 1.6\% on MolPCBA. Most of the improvement is attributed to the directional derivative aggregator, highlighting our method's ability of capturing directional high-frequency signals in graphs.

\section{Theoretical development}

\subsection{Intuitive overview}

One of the biggest limitations of current GNN methods compared to CNNs is the inability to do message passing in a specific direction such as the horizontal one in a grid graph. In fact, it is difficult to define directions or coordinates based solely on the shape of the graph. 

The lack of directions strongly limits the discriminative abilities of GNNs to understand local structures and simple feature transformations. 
Most GNNs are invariant to the permutation of the neighbours' features, so the nodes' received signal is not influenced by swapping the features of two neighbours. Therefore, several layers in a deep network will be employed to understand these simple changes instead of being used for higher level features, leading to problematic phenomena such as a  over-squashing \cite{alon_bottleneck_2020}. 


In the first part of the theoretical development, we develop the mathematical theory for general vector fields $\mF$. Intuitively, defining a vector field over a graph corresponds to assigning a scalar weight to edges corresponding to the magnitude of the flow in that direction. Note that $\mF$ has the same shape as the adjacency matrix and the same zero entries. As an example a left-to-right flow in a grid corresponds to a matrix with positive values over all left-to-right edges, negative over the right-to-left edges and 0 on the vertical edges.

In the second part, we set $\mF$ to be the gradient of the low-frequency eigenvectors of the Laplacian. Using this directional field, we show that the expressiveness of GNNs can be improved, while providing an intuitive directional flows over a variety of graphs (see figure \ref{fig:eig_vec_grad}). For example, we prove that in grid-shaped graphs some of these eigenvectors correspond to the horizontal and vertical flows. Again, we observe in the Minnesota map that the first 3 non-constant eigenvectors produce logical directions, namely South/North, suburb/city, and West/East.



Another important contribution---also noted in figure \ref{fig:eig_vec_grad}---is the ability to define any kind of directional flow based on prior knowledge of the problem. Hence, instead of relying on eigenvectors to find directions in a map, we can simply use the cardinal directions or the rush-hour traffic flow.

\subsection{Overview of the theoretical contributions}

\xhdr{Vector fields in a graph}
Using directions in a graph is novel and not intuitive, so our first step is to define a simple nomenclature where we use a vector field to define a directional flow at each node.

\xhdr{Directional smoothing and derivatives}
To make use of vector fields over graphs, we define aggregation matrices that can either smooth the signal (low pass filter) or compute its derivative (high pass filter) according to the directions specified by the vector field.

\xhdr{Gradient of the Laplacian eigenvectors} We show that using the gradient of the low-frequency eigenvectors of the graph Laplacian generates interpretable vector fields that counteract the over-smoothing problem.

\xhdr{Generalization of CNNs} We demonstrate that, when applied to a grid graph, the eigenvector-based directional aggregation generalizes convolutional neural networks.

\xhdr{Comparison to the Weisfeiler-Lehman (WL) test} We prove that the proposed DGN is more expressive than the 1-WL test, and thus more expressive than ordinary GNNs.

\subsection{Vector fields in a graph}
This section presents the ideas of differential geometry applied to graphs, with the goal of finding proper definitions of scalar products, gradients and directional derivatives. For reference see for example \cite{bronstein_geometric_2017,Grebenkov_2013,discrete_calculus_2010}.


Let $G=(V,E)$ be a graph with $V$ the set of vertices and $E\subset V\times V$ the set of edges. The graph is undirected meaning that $(i,j) \in E$ iff $(j,i) \in E$. Define the vector spaces $L^2(V)$ and $L^2(E)$ as the set of maps $V \to \R$ and $E \to \R$ with $\vx, \vy \in L^2(V)$ and $\mF, \mH \in L^2(E)$ and scalar products
\begin{equation}
\label{eq:dot_product}
    \begin{split}
    \langle \vx,\vy \rangle_{L^2(V)}: &= \sum_{i\in V} \vx_i \vy_i
    \\
    \langle \mF,\mH \rangle_{L^2(E)}: &= \sum_{(i,j)\in E} \mF_{(i,j)} \mH_{(i,j)}
    \end{split}
\end{equation}

Think of $E$ as the ``tangent space" to $V$ and of $L^2(E)$ as the set of ``vector fields'' on the space $V$ with each row $\mF_{i,:}$ representing a vector at the $i$-th node, and the element $\mF_{i,j}$ being the component of the vector going from node $i$ to $j$ through edge $e_{ij}$. Note that with $n$ the number of nodes in $G$, any $\vx \in L^2(V)$ can be represented as an $n$ coordinates vector and $\mF \in L^2(E)$ can be represented as an $n\times n$ matrix.

Define the pointwise scalar product as the map $L^2(E)\times L^2(E) \to L^2(V)$ taking 2 vector fields and returning their inner product at each point of $V$,  at the node $i$ is defined by \eqref{eq:inner-product}.

\begin{equation}
\label{eq:inner-product}
\langle \mF,\mH\rangle_{i}:=\sum_{j:(i,j)\in E}\mF_{i,j} \mH_{i,j}
\end{equation}

In \eqref{eq:grad_div}, we define the gradient $\nabla$ as a mapping $L^2(V) \to L^2(E)$ and the divergence $\diver$ as a mapping $L^2(E) \to L^2(V)$, thus leading to an analogue of the directional derivative in \eqref{eq:directional derivative}.
\begin{equation}
    \label{eq:grad_div}
    \begin{split}
    (\nabla \vx)_{(i,j)} &:= \vx(j)-\vx(i)
    \\
    (\diver \mF)_i &:= \sum_{j:(i,j) \in E} \mF_{(i,j)}
    \end{split}
\end{equation}

\begin{definition}
The directional derivative of the function $\vx$ on the graph $G$ in the direction of the vector field $\hat{\mF}$ where each vector is of unit-norm is 
\begin{equation}
\label{eq:directional derivative} 
D_{\hat{\mF}} \vx(i):=\langle \nabla \vx,\hat{\mF} \rangle_i =
\sum_{j:(i,j)\in E}(\vx(j)-\vx(i)) \hat{\mF}_{i,j}
\end{equation}
\end{definition}

$|\mF|$ will denote the absolute value of $\mF$ and $||\mF_{i,:}||_{L^p}$ the $L^p$-norm of the $i$-th row of $\mF$. We also define the forward/backward directions as the positive/negative parts of the field $\mF^\pm$.

\subsection{Directional smoothing and derivatives}
\label{sec:B_av-B_dx}

Next, we show how the vector field $\mF$ is used to \textit{guide} the graph aggregation by projecting the incoming messages. Specifically, we define the weighted aggregation matrices $\mB_{av}$ and $\mB_{dx}$ that allow to compute the directional smoothing and directional derivative of the node features, as presented visually in figure \ref{fig:full-method}-d.

\paragraph{The directional average matrix $\mB_{av}$}  is the weighted aggregation matrix such that all weights are positives and all rows have an $L^1$-norm equal to 1, as shown in \eqref{eq:Bav_simple} and theorem \ref{th:dir_smooth}, with a proof in the appendix \ref{app:proof:dir_smooth}.
\begin{equation}
\label{eq:Bav_simple}
\mB_{av}(\mF)_{i,:} = \frac{|\mF_{i,:}|}{||\mF_{i,:}||_{L^1} + \epsilon}
\end{equation}
The variable $\epsilon$ is an arbitrarily small positive number used to avoid floating-point errors.
The $L^1$-norm denominator is a local row-wise normalization.
The aggregator works by assigning a large weight to the elements in the forward or backward direction of the field, while assigning a small weight to the other elements, with a total weight of 1.

\begin{theorem}[Directional smoothing]
\label{th:dir_smooth}
    \Copy{th:dir_smooth}{
    The operation $\vy = \mB_{av} \vx$ is the directional average of $\vx$, in the sense that $\vy_u$ is the mean of $\vx_v$, weighted by the direction and amplitude of $\mF$.
    }
\end{theorem}

With $\vx_v$ the features at the nodes $v$ neighbouring $u$, and $\vy_u$ the directional smoothing at node $u$.

\paragraph{The directional derivative matrix $\mB_{dx}$} is defined in (\ref{eq:Bdx_simple}) and theorem \ref{th:dir_dx}, with the proof in appendix \ref{app:proof:dir_dx}. Again, the denominator is a local row-wise normalization but can be replaced by a global normalization. $\text{diag}(\va)$ is a square, diagonal matrix with diagonal entries given by $\va$. The aggregator works by subtracting the projected forward message by the backward message (similar to a center derivative), with an additional diagonal term to balance both directions.
\begin{equation}
\label{eq:Bdx_simple}
\begin{split}
\mB_{dx}(\mF)_{i,:} &= \hat{\mF}_{i,:} - \text{diag}\Big(\sum_{j}{\hat{\mF}_{:,j}} \Big)_{i,:}
\\
\hat{\mF}_{i,:} &= \left( \frac{\mF_{i,:}}{||\mF_{i,:}||_{L^1} + \epsilon}
\right)
\end{split}
\end{equation}

\begin{theorem}[Directional derivative]
\label{th:dir_dx}
    \Copy{th:dir_dx}{
    Suppose $\hat{\mF}$ have rows of unit $L^1$ norm. The operation $\vy = \mB_{dx}(\hat{\mF}) \vx$ is the centered directional derivative of $\vx$ in the direction of $\mF$, in the sense of equation \ref{eq:directional derivative}, i.e.
    \[\vy = D_{\hat{\mF}} \vx = 
    \Big( \hat{\mF} - \mathrm{diag} \Big(\sum_j \hat{\mF}_{:,j} \Big)\Big)\vx\]
    }
\end{theorem}

These aggregators are directional, interpretable and complementary, making them ideal choices for GNNs. We discuss the choice of aggregators in more details in appendix \ref{app:agg_choices}, while also providing alternative aggregation matrices such as the center-balanced smoothing, the forward-copy, the phantom zero-padding, and the hardening of the aggregators using softmax/argmax on the field. We further provide a visual interpretation of the $\mB_{av}$ and $\mB_{dx}$ aggregators in figure \ref{fig:directional-agg}. Interestingly, we also note in appendix \ref{app:simple_agg_examples} that $\mB_{av}$ and $\mB_{dx}$ yield respectively the mean and Laplacian aggregations when $\mF$ is a vector field such that all entries are constant $\mF_{ij}=\pm C$.

\begin{figure}[h]
\centering
 \includegraphics[width=\textwidth]
 {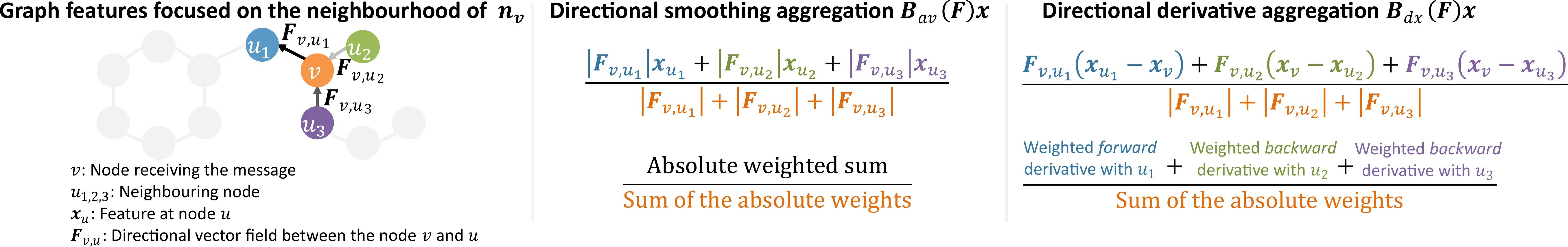}
 \vspace{-8pt}
 \caption{Illustration of how the directional aggregation works at a node $n_v$, with the arrows representing the direction and intensity of the field $\mF$.}
\label{fig:directional-agg}
\end{figure}

\subsection{Gradient of the Laplacian eigenvectors as interpretable vector fields}
\label{sec:grad_eig}

In this section we give theoretical support for the choice of gradients of the eigenfunctions of the Laplacian as sensible vectors along which to do directional message passing since they are interpretable and allow to reduce the over-smoothing. This section gives a theoretical ground to the intuitive directions presented in figure \ref{fig:eig_vec_grad}, and is the motivation behind steps (b-c) in figure \ref{fig:full-method}.

As usual the combinatorial, degree-normalized and symmetric normalized Laplacian are defined as
\begin{equation}
    \label{eq:laplacian}
    \mL = \mD - \mA
    ,\quad
    \mL_\text{norm} = \mD^{-1} \mL
    ,\quad
    \mL_\text{sym} = \mD^{-\frac{1}{2}} \mL \mD^{-\frac{1}{2}}
\end{equation}
The eigenvectors of these matrices are known to capture many essential properties of graphs, making them a natural foundation for directional message passing.
For example, the Laplacian eigenvectors corresponding to the smallest eigenvalues (i.e., the low frequency eigenvectors) effectively capture the community structure of a graph, and these eigenvectors also play the role of Fourier modes in graph signal processing \cite{hamilton_2020}.
Indeed, the Laplacian eigenvectors hold such rich information about graph structure that their study is the focus of the mathematical subfield of spectral graph theory
\cite{chung1997spectral}.

In order to illustrate the utility of these eigenvectors in the context of GNNs, we show that the low-frequency eigenvectors provide a natural direction that allows us to pass messages between distant nodes in a graph. 
In particular, we show in theorem \ref{th:reduced-diffusion-distance} (proved in appendix \ref{app:proof:reduced-diffusion-distance}) that by passing information in the direction of $\eigvec_1$, the eigenvector associated to the lowest non-trivial frequency of $\mL_\text{norm}$, DGNs can efficiently share information between distant nodes of the graph by reducing the diffusion distance between them. This idea is reflected in figure \ref{fig:eig_vec_grad}, where we see that the eigenvectors of the Laplacian give directions that correspond to a natural notion of distance on real-world graphs.


In the next paragraphs, we will prove that following the gradient of the eigenvectors allows to effectively reduce the heat-kernel distance between pairs of nodes.

Consider the transition matrix $\mW=\mD^{-1}\mA$. Its entries can be used to define a random walk with probability to move from node $x$ to node $y$ equal to $p_1(x,y) = \frac{1}{d_x}$ if $x$ and $y$ are neighbors and $0$ if not. Notice that the probability to transition from $x$ to $y$ in $k$ steps is given by the $x,y$ entry of the matrix $\mW^k$. This matrix is also called the discrete heat kernel $p_k(x,y)= (\mW^k)_{x,y}$. 
Given a Markov process $\Tilde{X}_k$ defined by the transition matrices $\mW^k$, $j=1,...,k$, we can define a continuous time random walk on the same graph in the following way. Let $N_t$ be a mean 1 Poisson random variable, the continuous time random variable is defined by $X_t := \Tilde{X}_{N_t}$ with transition probability $q_t(x,y) = P(X_t=y|x_0=x)$.

In \cite{barlow_2017}, the following identity is shown 
\[q_t(x,y) = \sum_{n=0}^{\infty} \frac{e^{-t}t^k}{k!}p_k(x,y)
\]
Or in matrix form $q_t = e^{t(\mW - \mI)} = e^{-t\mL_{\text{norm}}}$. This transition probability is also called the continuous time heat kernel because it satisfies the continuous time heat equation on graphs $\frac{d}{dt}q_t = -\mL_{\text{norm}}q_t$. In \cite{COIFMAN20065} the following distance is defined
\begin{definition}[Diffusion distance]
The diffusion distance at time $t$ between the nodes $x,y$ is 
\begin{equation}
    d_t(x,y):=\left( \sum_{z\in V} \Big(q_t(x,z)-q_t(y,z)\Big)^2 \right)^\frac{1}{2}
\end{equation}
\end{definition}
The diffusion distance is small when there is high probability that two random walks starting at $x$ and $y$ meet at time $t$. The diffusion distance is used as a model of how the data at a node $x$ influences a node $y$ in a GNN. The symmetrisation of the heat kernel in the diffusion distance and the use of continuous time are slight departure from the actual process of information diffusion in a GNN but allow us to describe the important phenomenons with much simpler statements.

\begin{definition}[Gradient step]
\label{def:gradient-step}
\Copy{def:gradient-step}{Suppose the two neighboring nodes $x$ and $z$ are such that $\eigvec(z) - \eigvec (x)$ is maximal among the neighbors of $x$, then we will say $z$ is obtained from $x$ by taking a step in the direction of the gradient $\nabla \eigvec$.
}
\end{definition}
\begin{theorem}[Gradient steps reduce diffusion distance]
\label{th:reduced-diffusion-distance}
\Copy{th:reduced-diffusion-distance}{
Let $x,y$ be nodes such that $\eigvec_1(x)<\eigvec_1(y)$. Let $x'$ be the node obtained from $x$ by taking one step in the direction of $\nabla \eigvec_1$, then there is a constant $C$ such that for $ C \leq t$ we have 
\[d_t(x',y) < d_t(x,y).
\]
With the reduction in distance being proportional to $e^{-\lambda_1}$.
}
\end{theorem}
From this theorem, we see that moving from node $x$ to node $x'$ by following the gradient of the eigenvector $\eigvec_1$ is guaranteed to reduce the heat kernel distance with a destination node $y$.
While the theorem always holds for $\eigvec_1$, it should be true for higher frequency eigenvectors if the graph has added structure for example if it is an approximation of a surface or a higher dimensional manifold.

In the context of GNNs, Theorem \ref{th:reduced-diffusion-distance} also has implications for the well-known problems of {\em over-smoothing} and {\em over-squashing} \cite{alon_bottleneck_2020, hamilton_2020}. 
In most GNN models, node representations become over-smoothed after several rounds of message passing, as the representations tend to reach a mean-field equilibrium equivalent to the stationary distribution of a random walk \cite{hamilton_2020}. 
Researchers have also highlighted the related issue of over-squashing, which reflects the inability for GNNs to propagate informative signals between distant nodes in a graph \cite{alon_bottleneck_2020}.

Both these problems are related to the fact that the influence of one node's input on the final representation of another node in a GNN is correlated with the diffusion distance between the nodes \cite{xu2018representation}.
Theorem \ref{th:reduced-diffusion-distance} highlights how the DGN approach can alleviate these issues. In particular, the Laplacian eigenfunctions reveal directions that can counteract over-smoothing and over-squashing by allowing efficient propagation of information between distant nodes instead of following a diffusion process. 

Finally it is interesting to note that by selecting different eigenvectors as basis of directions, our method further aligns with a theorem that multiple independent aggregators are needed to distinguish neighbourhoods of nodes with continuous features \cite{corso2020principal}.

\subsection{Choosing a basis of the Laplacian eigenspace}

When using eigenvectors of the Laplacian $\eigvec_i$ to define directions in a graph, we need to keep in mind that there is never a single eigenvector associated to an eigenvalue, but a whole eigenspace.
If an eigenvalue has multiplicity of $k$, the associated eigenspace has dimension $k$ and any collection of $k$ orthogonal vectors could be chosen as basis of that space and as vectors for the definitions of the aggregation matrices $\mB$ defined in the previous sections.

\xhdr{Disconnected graphs} When a graph is disconnected, then the eigenfunctions will simply be the combination of the eigenfunctions of each connected components. Hence, one must consider $\eigvec_i$ as the $i$-th eigenvector of each component when taken separately.

\xhdr{Normalizing the eigenvectors} For an eigenvalue of multiplicity 1, there are always two unit norm eigenvectors of opposite sign, which poses a problem during the directional aggregation. We can make a choice of sign and later take the absolute value (i.e. $\mB_{av}$ in \eqref{eq:Bav_simple}).
An alternative that applies to multiplicities higher than 1 is to take samples of orthonormal bases of the eigenspace and use each choice to augment the training (see section \ref{sec:data-augmentation}). 

\xhdr{Multiplicities greater than 1} Although multiplicities higher than one do happen for low-frequencies (square grids have a multiplicity 2 for $\lambda_1$) this is not common in ``real-world graphs'' since it suggests symmetries in the graph which are uncommon. Furthermore,  we found no $\lambda_1$ multiplicity greater than 1 in the ZINC and PATTERN datasets. We further discuss these rare cases and how to deal with them in appendix \ref{app:eig_mult}.

\xhdr{Orthogonal directions} Although all $\eigvec$ are orthogonal, their gradients, used to define directions, are not always \textit{locally} orthogonal (e.g. there are many horizontal flows in the grid).
This concern is left to be addressed in future work.

\subsection{Generalization of the convolution on a grid} \label{sec:generalization_cnn}

In this section we show that our method generalizes CNNs by allowing to define any radius-$R$ convolutional kernels in grid-shaped graphs. The radius-$R$ kernel at node $u$ is a convolutional kernel that takes the weighted sum of all nodes $v$ at a distance $d(u,v) \le R$.

Consider the lattice graph $\Gamma$ of size $N_1 \times N_2 \times ... \times N_n$ where each vertices are connected to their direct non-diagonal neighbour. We know from Lemma \ref{lemma:cosine_eigvec} that, for each dimension, there is an eigenvector that is only a function of this specific dimension. For example, the lowest frequency eigenvector $\eigvec_1$ always flows in the direction of the longest length. Hence, the Laplacian eigenvectors of the grid can play a role analogous to the axes in Euclidean space, as shown in figure \ref{fig:eig_vec_grad}. 

With this knowledge, we show in theorem \ref{th:general_grid_radius2} (proven in \ref{app:proof:general_grid_radius2}), that we can generalize all convolutional kernels in an n-dimensional grid.
This is a strong result since it demonstrates that our DGN framework generalizes CNNs when applied on a grid, thus closing the gap between GNNs and the highly successful CNNs on image tasks.

\begin{theorem}[Generalization radius-$R$ convolutional kernel in a lattice]
\label{th:general_grid_radius2}
    \Copy{th:general_grid_radius2}{
    For an $n$-dimensional lattice, any convolutional kernel of radius $R$ can be realized by a linear combination of directional aggregation matrices and their compositions.
}
\end{theorem}

\begin{figure}[h]
    \centering
    \includegraphics[width=\textwidth]{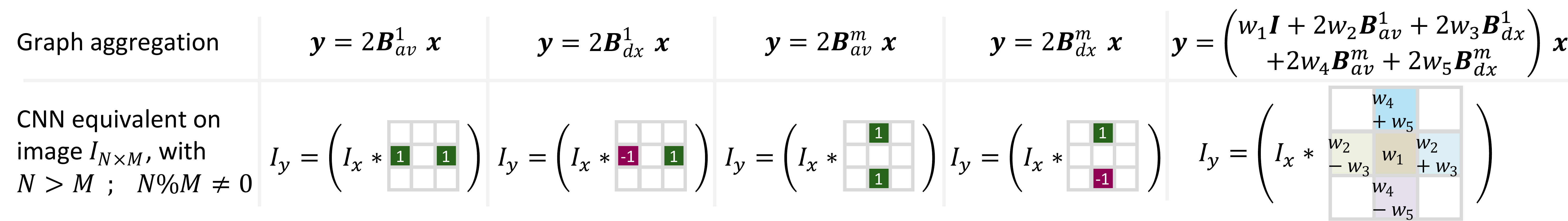}
    \vspace{-10pt}
    \caption{Realization of a radius-1 convolution using the proposed aggregators. $I_x$ is the input feature map, $*$ the convolutional operator, $I_y$ the convolution result, and $\mB^i = \mB(\nabla \eigvec_i)$.}
    \label{fig:cnn-gnn-generalization}
\end{figure}

As an example, figure \ref{fig:cnn-gnn-generalization} shows how a linear combination of the first and $m$-th aggregators $\mB(\nabla \eigvec_{1,m})$ realize a kernel on an $N\times M $ grid, where $m = \ceil{N/M}$ and $N>M$.

Note that when the size of a given dimension is an integer multiple of another direction, e.g. $N=M$ or $N=3 M$, then you will find a multiplicity of 2 for the $m-th$ eigenvector. Hence, the eigenvector used to define the direction is not unique. This does not void theorem \ref{th:general_grid_radius2} since the eigenvectors flowing in the horizontal/vertical directions are still valid choices.



\subsection{Extending the radius of the aggregation kernel}

Having aggregation kernels for neighbours of distance 2 or 3 is important to improve the expressiveness of GNNs, their ability to understand patterns, and to reduce the number of layers required. However, the lack of directions in GNNs strongly limits the radius of the kernels since, given a graph of regular degree $d$, a mean/sum aggregation at a radius-$R$ will result in a heavy over-squashing of $O(d^R)$ messages. Using the directional fields, we can enumerate different paths, thus assigning a different weight for different $R$-distant neighbours. This method, proposed in appendix \ref{app:radius-r-kernel}, avoids the over-squashing. (Empirical results on this extension are left for future work.)

\subsection{Comparison with Weisfeiler-Lehman (WL) test}

We also compare the expressiveness of the Directional Graph Networks with the classical WL graph isomorphism test which is often used to classify the expressivity of graph neural networks \cite{xu2018gin}. In theorem \ref{th:wl-test} (proven in appendix \ref{app:proof:wl-test}) we show that DGNs are capable of distinguishing pairs of graphs that the 1-WL test (and so ordinary GNNs) cannot differentiate.

\begin{theorem}[Comparison with 1-WL test]
\label{th:wl-test}
    \Copy{th:wl-test}{
    DGNs using the \textit{mean} aggregator, any directional aggregator of the first Laplacian eigenvector and injective degree-scalers are strictly more powerful than the 1-WL test.
}
\end{theorem}

\subsection{Data augmentation}\label{sec:data-augmentation}

Another theoretical result is that the directions in the graph allow to replicate some of the most common data augmentation techniques used in computer vision, namely reflection, rotation and distortion. The main difference is that, instead of modifying the image (such as a $5^\circ$ rotation), the proposed transformation is applied on the vector field defining the aggregation kernel (thus rotating the kernel by $-5^\circ$ without changing the image). This offers the advantage of avoiding to pre-process the data since the augmentation is done directly on the kernel at each iteration of the training.

The simplest augmentation is the vector field flipping, which is done changing the sign of the field $\mF$, as stated in definition \ref{def:field_flip}. This changes the sign of $\mB_{dx}$, but leaves $\mB_{av}$ unchanged.

 \begin{definition}[Reflection of the vector field]
\label{def:field_flip}
For a vector field $\mF$, the reflected field is $-\mF$.
\end{definition}

Let  $\mF_1, \mF_2$ be vector fields in a graph, with $\hat{\mF}_1$ and $\hat{\mF}_2$ being the field normalized such that each row has a unitary $L^2$-norm. Define the angle vector $\bm{\alpha}$ by $\langle (\hat{\mF}_1)_{i,:},(\hat{\mF}_2)_{i,:}\rangle = \cos(\bm{\alpha}_i)$. The vector field $\hat{\mF}^\perp_2$ is the normalized component of $\hat{\mF}_2$ perpendicular to $\hat{\mF}_1$. The equation below defines $\hat{\mF}^\perp_2$. The next equation defines the angle 
\[
    (\hat{\mF}_2^\perp)_{i,:} = \frac{
    (\hat{\mF}_2 - \langle \hat{\mF}_1, \hat{\mF}_2 \rangle \hat{\mF}_1)_{i,:}}
    {||(\hat{\mF}_2 - \langle \hat{\mF}_1, \hat{\mF}_2 \rangle \hat{\mF}_1)_{i,:}||}
\]

Notice that we then have the decomposition $(\hat{\mF}_2)_{i,:} = \cos(\bm{\alpha}_i)(\hat{\mF}_1)_{i,:} + \sin(\bm{\alpha}_i)(\hat{\mF}_2^\perp)_{i,:}$.

\begin{definition}[Rotation of the vector fields]
\label{def:field_rotation}
    \Copy{def:field_rotation}{
    For $\hat{\mF}_1$ and $\hat{\mF}_2$ non-colinear vector fields with each vector of unitary length, their rotation by the angle $\theta$ in the plane formed by $\{\hat{\mF}_1, \hat{\mF}_2\}$ is     
    \begin{equation}
        \label{eq:field_rotation}
        \begin{split}
        \hat{\mF}^\theta_1 &= \hat{\mF}_1 \textnormal{diag} (\cos \theta) + \hat{\mF}^\perp_2 \textnormal{diag} (\sin \theta)
        \\
        \hat{\mF}^\theta_2 &= \hat{\mF}_1 \textnormal{diag} (\cos (\theta + \bm{\alpha})) + \hat{\mF}^\perp_2  \textnormal{diag}(\sin (\theta + \bm{\alpha}))
        \end{split}
    \end{equation}
}
\end{definition}

Finally, the following augmentation has a similar effect to a wave distortion applied on images.

\begin{definition}[Random distortion of the vector field]
\label{def:field_distortion}
    \Copy{def:field_distortion}{
    For vector field $\mF$ and anti-symmetric random noise matrix $\mR$, its randomly distorted field is $\mF' = \mF + \mR \circ \mA$.
    }
\end{definition}

\section{Implementation}
\label{sec:implementation}

We implemented the models using the DGL and PyTorch libraries and we provide the code at the address
\href{https://github.com/Saro00/DGN}{https://github.com/Saro00/DGN}. 
We test our method on standard benchmarks from \cite{dwivedi2020benchmarking} and \cite{hu2020open}, namely ZINC, CIFAR10, PATTERN, MolHIV and MolPCBA with more details on the datasets and how we enforce a fair comparison in appendix \ref{app:benchmarks}.

For the empirical experiments we inserted our proposed aggregation method in two different type of message passing architectures used in the literature: a \textit{simple} convolutional architecture similar to the one present in GCN (equation \ref{eq:simple}) \cite{kipf2016gcn} and a more \textit{complex} and general one typical of MPNNs (\ref{eq:complex}) \cite{gilmer2017mpnn} with or without edge features $e_{ji}$. The time complexity of our approach is $O(Em)$, which is identical to PNA \cite{corso2020principal}, where $E$ is the number of edges and $m$ the number of aggregators, with an additional $O(Ek)$ to pre-compute the $k$-first eigenvectors, as explained in the appendix \ref{app:computation-complexity}.
\begin{subequations}
  \begin{equation}
     \label{eq:simple}
    X_i^{(t+1)} = 
    U \Bigg(
    \underset{(j,i) \in E}{\bigoplus} 
    X_j^{(t)} \Bigg)
  \end{equation}
  \begin{equation}
    \label{eq:complex}
    X_i^{(t+1)} = 
    U \Bigg( X_i^{(t)}, 
    \underset{(j,i) \in E}{\bigoplus} 
    M \Big( X_i^{(t)}, X_j^{(t)}, 
    \underbrace{e_{ji}}_{\text{\tiny optional}}
    \Big) \Bigg)
  \end{equation}
\end{subequations}
Here, $\bigoplus$ is an operator which concatenates the results of multiple aggregators, $X$ is the node features, $M$ is a linear transformation and $U$ a multiple layer perceptron (MLP). This \textit{simple} architecture of equation \ref{eq:simple} is observed visually in steps (f-g) of figure \ref{fig:full-method}.

We further use degree scalers $S(d, \alpha)$ defined below to scale the aggregation results according to each node's degree, as proposed by the PNA model \cite{corso2020principal}. Here, $d$ is the degree of a given node, $\delta$ is the average node degree in the training set, and $\alpha$ is a parameter set to $-1$ for degree-attenuation and $1$ for degree amplification. Note that each degree scaler is applied to the result of each aggregator, and the results are concatenated.
\begin{equation}
\label{eq:Scaler}
S(d, \alpha) = \left(\frac{\log(d + 1)}{\delta} \right)^\alpha, \; \delta = \frac{1}{|\text{train}|}\sum_{i \, \in \,  \text{train}}\log(d_i + 1)
\end{equation}

We tested the directional aggregators across the datasets using the gradient of the first $k$ eigenvectors $\nabla \eigvec_{1,...,k}$ as the underlying vector fields. Here, $k$ is a hyperparameter, usually 1 or 2, but could be bigger for high-dimensional graphs. To deal with the arbitrary sign of the eigenvectors, we take the absolute value of the result of \eqref{eq:Bdx_simple}, making it invariant to a reflection of the field. In case of a disconnected graph, $\eigvec_i$ is the $i$-th eigenvector of each connected component. Despite the numerous aggregators proposed in appendix \ref{app:agg_choices}, only $\mB_{dx}$ and $\mB_{av}$ are tested empirically.

The metrics used to measure the performance of a model depend are enforced for each dataset and provided by \cite{dwivedi2020benchmarking} and \cite{hu2020open}. In particular, we use the mean absolute error (MAE), the accuracy (acc), the area under the receiver operating curve (ROC-AUC), and the average precision (AP).

\section{Results and discussion}

\subsection{Directional aggregation}

Using the benchmarks introduced in section \ref{sec:implementation}, we present in figure \ref{fig:results_table} a fair comparison of various aggregation strategies using the same parameter budget and hyperparameters. We see a consistent boost in the performance for \textit{simple}, \textit{complex} and \textit{complex with edges} models using directional aggregators compared to the \textit{mean-aggregator} baseline.

With our theoretical analysis in mind, we expected to perform well on PATTERN since the flow of the first eigenvectors are meaningful directions in a stochastic block model (i.e., these eigenvectors tend to correlate with community membership).
The results match our expectations, outperforming all the previous models.

In particular, we see a significant improvement in the molecular datasets (ZINC, MolHIV and MolPCBA) when using the directional aggregators, especially for the derivative aggregation $\mB_{dx}^1$ (noted \textit{dx\textsubscript{1}} in figure \ref{fig:results_table}). 
We believe this is due to the capacity to efficiently move  messages  across opposite parts of the molecule and to better understand the role of atom pairs. We further believe that the derivative aggregator is better able to capture high-frequency directional signals, similarly to the Gabor filters in computer vision.

Further, the thesis that DGNs can bridge the gap between CNNs and GNNs is supported by the clear improvements on CIFAR10 over the baselines. 

In the work by \cite{dwivedi2020benchmarking}, they proposed the use of positional encoding of the eigenvectors. However, our experiments with the positional encoding of the first 2 non-trivial eigenvectors, noted \textit{pos\textsubscript{1}, pos\textsubscript{2}} in figure \ref{fig:results_table}, showed no clear improvement on most datasets. In fact, Dwivedi et al. noted that many eigenvectors and high network depths are required for improvements, yet we outperform their results with fewer parameters, less depth, and only 1-2 eigenvectors, further motivating their use as directional flows instead of positional encoding.

\subsection{Comparison to the literature}

\begin{figure*}[ht]
\centering
\includegraphics[height=4.3cm]{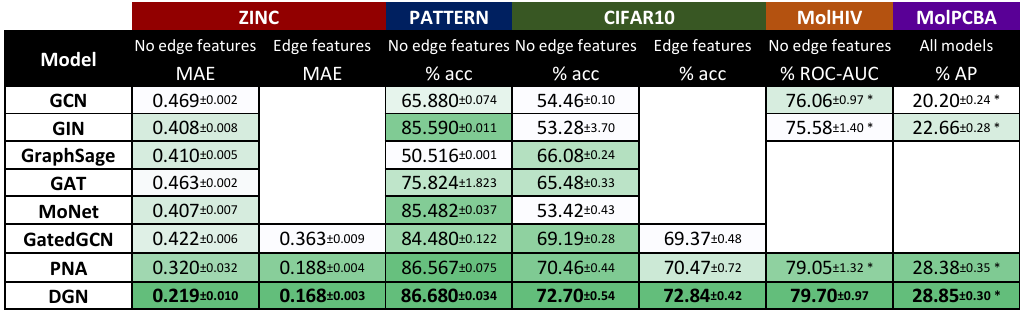}
\vspace{-8pt}
\caption{Fine-tuned results of the DGN model against models from \cite{dwivedi2020benchmarking} and \cite{hu2020open}: GCN \cite{kipf2016gcn}, GraphSage \cite{hamilton2017inductive}, GIN \cite{xu2018gin}, GAT \cite{velikovic2017gat}, MoNet \cite{monti2017moNet}, GatedGCN \cite{bresson2017gatedGCN} and PNA \cite{corso2020principal}. All the models use $\sim100k$ parameters, except those with * who use $300k$ to $6.5M$. In ZINC the DGN aggregators are \textit{\{mean, dx$_1$, max, min\}}, in PATTERN \textit{\{mean, dx$_1$, av$_1$\}}, in CIFAR10 \textit{\{mean, dx$_1$, dx$_2$, max\}}, in MolHIV \textit{\{mean, dx$_1$, av$_1$, max, min\}}, in MolPCBA \textit{\{mean, sum, max, dx$_1$\}}. Mean and uncertainty are taken over 4 runs for ZINC, PATTERN and CIFAR10 and 10 runs for MolHIV and MolPCBA.
}
\label{fig:resultscomparison}
\end{figure*}

In order to compare our model with the literature, we fine-tuned it on the various datasets and we report its performance in figure \ref{fig:resultscomparison}. We observe that DGN provides significant improvement across all benchmarks, highlighting the importance of anisotropic kernels that are dependant on the graph topology.

\begin{figure*}[ht]
\centering
\includegraphics[height=4.3cm]{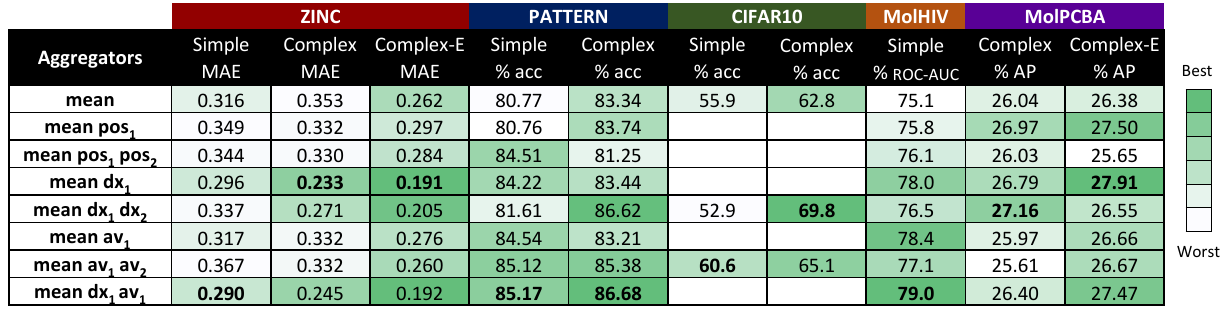}
\vspace{-8pt}
\caption{Test set results using a parameter budget of $\sim100k$ with the same hyperparameters as \cite{corso2020principal}, except MolPCBA with a budget of $\sim7M$. The low-frequency Laplacian eigenvectors are used to define the directions, except for CIFAR10 that uses the coordinates of the image. For brevity, we denote \textit{dx}\textsubscript{i} and \textit{av}\textsubscript{i} as the directional derivative $\mB_{dx}^i$ and smoothing $\mB_{av}^i$ aggregators of the $i$-th direction. We also denote \textit{pos}\textsubscript{i} as the $i$-th eigenvector used as positional encoding for the mean aggregator.}
\label{fig:results_table}
\end{figure*}

Note that the results in Figure \ref{fig:resultscomparison} are better those in Figure \ref{fig:results_table} since the latter uses a more exhaustive parameter search, and uses the \textit{min/max} aggregators proposed in PNA \cite{corso2020principal} alongside the directional aggregators.

\subsection{Preliminary results of data augmentation}

\begin{figure*}[h]
\centering
\includegraphics[width=\textwidth]{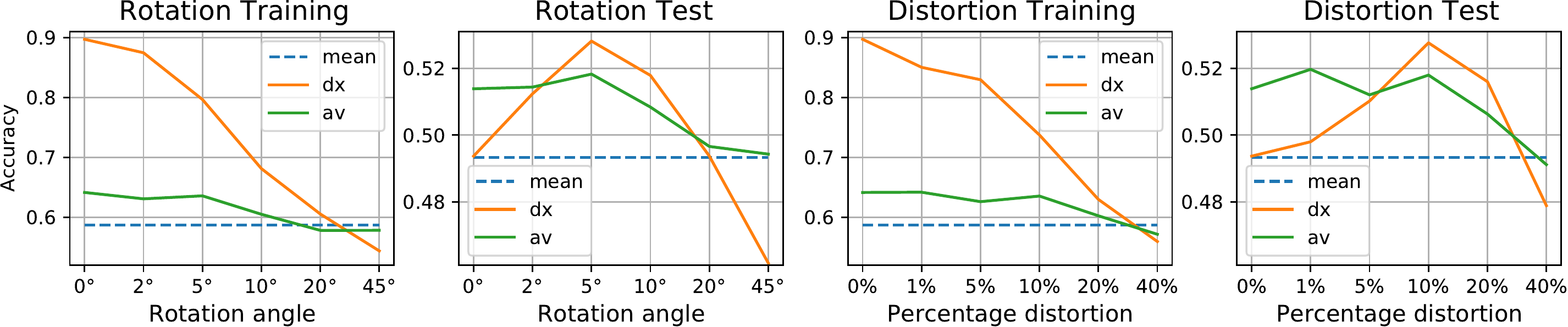}
\vspace{-18pt}
\caption{ Accuracy of the various models using data augmentation with a \textit{complex} architecture of $\sim100k$ parameters and trained on 10\% of the CIFAR10 training set (4.5k images). An angle of $x$ corresponds to a rotation of the kernel by a random angle sampled uniformly in $(-x \si{\degree}, x \si{\degree})$ using definition \ref{def:field_rotation} with $\mF_{1,2}$ being the gradient of the horizontal/vertical coordinates. A noise of $100x\%$ corresponds to a distortion of each eigenvector with a random noise uniformly sampled in $(-x \cdot m, x \cdot m)$ where $m$ is the average absolute value of the eigenvector's components. The \textit{mean} baseline model is not affected by the augmentation since it does not use the underlining vector field. }
\label{fig:results_augmentation}
\end{figure*}

To evaluate the effectiveness of the proposed augmentation, we trained the models on a reduced version of the CIFAR10 dataset. 
The results in figure \ref{fig:results_augmentation} show clearly a higher expressive power of the \textit{dx} aggregator, enabling it to fit well the training data. For a small dataset, this comes at the cost of overfitting and a reduced test-set performance, but we observe that randomly rotating or distorting the kernels counteracts the overfitting and improves the generalization. 

As expected, the performance decreases when the rotation or distortion is too high since the augmented graph changes too much. In computer vision images similar to CIFAR10 are usually rotated by less than $30^\circ$ \cite{shorten_survey_2019, ogara_comparing_2019}. Further, due to the constant number of parameters across models, less parameters are attributed to the mean aggregation in the directional models, thus it cannot fit well the data when the rotation/distortion is too strong since the directions are less informative. We expect large models to perform better at high angles.

\section{Conclusion}

The proposed DGN method allows to address many problems of GNNs, including the lack of anisotropy, the low expressiveness, the over-smoothing and over-squashing. For the first time in graph networks, we generalize the directional properties of CNNs and their data augmentation capabilities. Based on the intuitive idea that the low-frequency eigenvectors of the graph Laplacian gives an interpretable directional flow, we backed our work by a set of strong theoretical results showing that these eigenvectors are important in connecting nodes that are far away and improving the expressiveness in regards to the WL-test. 

The work being also supported by strong empirical results, we believe it will give rise to a new family of directional GNNs. In fact, we introduce in the appendix different avenues for future work, including the hardening of the aggregators \ref{app:agg:hardening}, the introduction of a zero-padding at the boundaries \ref{app:agg:zero-padding}, the implementation of radius-$R$ kernels \ref{app:radius-r-kernel}, and the full study of directional data augmentation. Future methods could also improve the choice of multiple directions beyond the selection of the \textit{k-lowest} frequencies.

\xhdr{Broader Impact}
This work will extend the usability of graph networks to all problems with engineering and physically defined directions, thus making GNN a new laboratory for signal processing, physics, material science and molecular and cell biology.
In fact, the anisotropy present in a wide variety of systems could be expressed as vector fields (spinor, tensor) compatible with the DGN framework, without the need of eigenvectors. One example is magnetic anisotropicity in metals, alloys and organic molecules that is dependant on the relative orientation to the magnetic field.
Other examples are the response of materials to high electromagnetic fields; all kind of field propagation in crystals lattices (vibrations, heat, shear and frictional force, young modulus, light refraction, birefringence); multi-body or liquid motion; magnons and solitons in different media, fracture propagation, traffic modelling; developmental biology and embryology, and design of novel materials and constrained structures. Finally applications based on neural operators for ODE/PDE may benefit as well.





\bibliography{citations_arxiv}
\bibliographystyle{plain}

\clearpage
\newpage
\onecolumn

\appendix

\title{Directional Graph Networks}
\renewcommand{\undertitle}{
Anisotropic aggregation in graph neural networks via directional vector fields
}

\section{Appendix - Choices of directional aggregators}
\label{app:agg_choices}

This appendix helps understand the choice of $\mB_{av}$ and $\mB_{dx}$ in section \ref{sec:B_av-B_dx} and presents different directional aggregators that can be used as an alternative to the ones proposed. 

A simple alternative to the directional smoothing and directional derivative operator is to simply take the \textit{forward/backward} values according to the underlying positive/negative parts of the field $\mF$, since it can effectively replicate them. However, there are many advantage of using $\mB_{av,dx}$. First, one can decide to use either of them and still have an interpretable aggregation with half the parameters. Then, we also notice that $\mB_{av,dx}$ regularize the parameter by forcing the network to take both forward and backward neighbours into account at each time, and avoids one of the neighbours becoming too important. Lastly, they are robust to a change of sign of the eigenvectors since $\mB_{av}$ is sign invariant and $\mB_{dx}$ will only change the sign of the results, which is not the case for \textit{forward/backward} aggregations.

\subsection{Retrieving the mean and Laplacian aggregations}
\label{app:simple_agg_examples}
It is interesting to note that we can recover simple aggregators from the aggregation matrices $\mB_{av}(\mF)$ and $\mB_{dx}(\mF)$. Let $\mF$ be a vector field such that all edges are equally weighted $\mF_{ij} = \pm C$ for all edges $(i,j)$. Then, the aggregator $\mB_{av}$ is equivalent to a mean aggregation:
\[\mB_{av}(\mF) \vx = \mD^{-1} \mA \vx
\]
Under the condition $F_{ij}=C$, the differential aggregator is equivalent to a Laplacian operator $\mL$ normalized using the degree $\mD$
\[\mB_{dx}(C \mA)\vx 
= \mD^{-1} (\mA - \mD) \vx 
= -\mD^{-1}  \mL \vx
\]

\subsection{Global field normalization}

The proposed aggregators are defined with a row-wise normalized field 
\[\hat{\mF_{i,:}} = \frac{\mF_{i,:}}{||\mF_{i,:}||_{L^P}}
\]
meaning that all the vectors are of unit-norm and the aggregation/message passing is done only according to the direction of the vectors, not their amplitude. However, it is also possible to do a global normalization of the field $\mF$ by taking a matrix-norm instead of a vector-norm. Doing so will modulate the aggregation by the amplitude of the field at each node. One needs to be careful since a global normalization might be very sensitive to the number of nodes in the graph.

\subsection{Center-balanced aggregators}

A problem arises in the aggregators $\mB_{dx}$ and $\mB_{av}$ proposed in equations \ref{eq:Bav_simple} and \ref{eq:Bdx_simple} when there is an imbalance between the positive and negative terms of $\mF^\pm$. In that case, one of the directions overtakes the other in terms of associated weights.

An alternative is also to normalize the forward and backward directions separately, to avoid having either the backward or forward direction dominating the message.
\begin{equation}
\mB_{av-center}(\mF)_{i,:} = 
\frac{\mF'^+_{i,:} + \mF'^-_{i,:}}{||\mF'^+_{i,j} + \mF'^-_{i,j}||_{L_1}}
\quad,\quad\quad
\mF'^\pm_{i,:} = \frac{|\mF^\pm_{i,:}|}{||\mF^\pm_{i,:}||_{L^1} + \epsilon}
\end{equation}

The same idea can be applied to the derivative aggregator \eqref{eq:Bdx_center} where the positive and negative parts of the field $\mF^\pm$ are normalized separately to allow to project both the \textit{forward} and \textit{backward} messages into a vector field of unit-norm. $\mF^+$ is the out-going field at each node and is used for the \textit{forward} direction, while $\mF^-$ is the in-going field used for the \textit{backward} direction. By averaging the \textit{forward} and \textit{backward} derivatives, the proposed matrix $\mB_{dx\text{-center}}$ represents the centered derivative matrix.

\begin{equation}
\label{eq:Bdx_center}
\mB_{dx\text{-center}}(\mF)_{i,:} = \mF_{i,:}' - \text{diag}\left(\sum_{j}{\mF'_{:,j}} \right )_{i,:}
\space,\quad
\mF_{i,:}' = \frac{1}{2} \left(\underbrace{ \frac{\mF_{i,:}^+}{||\mF_{i,:}^+||_{L^1} + \epsilon}}_{\text{forward field}}  + \underbrace{\frac{\mF_{i,:}^-}{||\mF_{i,:}^-||_{L^1} + \epsilon}}_{\text{backward field}}
\right)
\end{equation}

\subsection{Hardening the aggregators}
\label{app:agg:hardening}

The aggregation matrices that we proposed, mainly $\mB_{dx}$ and $\mB_{av}$ depend on a smooth vector field $\mF$. At any given node, the aggregation will take a weighted sum of the neighbours in relation to the direction of $\mF$. Hence, if the field $\mF_v$ at a node $v$ is \textit{diagonal} in the sense that it gives a non-zero weight to many neighbours, then the aggregator will compute a weighted average of the neighbours. 

Although there are clearly good reasons to have this weighted-average behaviour, it is not necessarily desired in every problem. For example, if we want to move a single node across the graph, this behaviour will smooth the node at every step. Instead, we propose below to soften and harden the aggregations by forcing the field into making a decision on the direction it takes.

\paragraph{Soft hardening the aggregation} is possible by using a softmax with a temperature $T$ on each row to obtain the field $\mF_\text{softhard}$.
\begin{equation}
    (\mF_\text{softhard})_{i,:} = \text{sign}(\mF_{i,:}) \space
    \text{softmax}(T |\mF_{i,:}|)
\end{equation}

\paragraph{Hardening the aggregation} is possible by using an infinite temperature, which changes the softmax functions into argmax. In this specific case, the node with the highest component of the field will be copied, while all other nodes will be ignored.

\begin{equation}
    (\mF_\text{hard})_{i,:} =
    \text{sign}(\mF_{i,:}) \space
    \text{argmax}(|\mF_{i,:}|)
\end{equation}

An alternative to the aggregators above is to take the \textit{softmin/argmin} of the negative part and the \textit{softmax/argmax} of the positive part.

\subsection{Forward and backward copy}

The aggregation matrices $\mB_{av}$ and $\mB_{dx}$ have the nice property that if the field is flipped (change of sign), the aggregation gives the same result, except for the sign of $\mB_{dx}$. However, there are cases where we want to propagate information in the forward direction of the field, without smoothing it with the backward direction. In this case, we can define the strictly forward and strictly backward fields below, and use them directly with the aggregation matrices.

\begin{equation}
    \mF_\text{forward} = \mF^+
    \quad,\quad\quad
    \mF_\text{backward} = \mF^-
\end{equation}

Further, we can use the hardened fields in order to define a forward copy and backward copy, which will simply copy the node in the direction of the highest field component.

\begin{equation}
    \mF_\text{forward copy} = \mF^+_\text{hard}
    \quad,\quad\quad
    \mF_\text{backward copy} = \mF^-_\text{hard}
\end{equation}

\subsection{Phantom zero-padding}
\label{app:agg:zero-padding}

Some recent work in computer vision has shown the importance of zero-padding to improve CNNs by allowing the network to understand it's position relative to the border \cite{islam_how_2020}. In contrast, using boundary conditions or reflection padding makes the network completely blind to positional information. In this section, we show that we can mimic the zero-padding in the direction of the field $\mF$ for both aggregation matrices $\mB_{av}$ and $\mB_{dx}$.

Starting with the $\mB_{av}$ matrix, in the case of a missing neighbour in the forward/backward direction, the matrix will compensate by adding more weights to the other direction, due to the denominator which performs a normalization. Instead, we would need the matrix to consider both directions separately so that a missing direction would result in zero padding. Hence, we define $\mB_{av,0pad}$ below, where either the $F^+$ or $F^-$ will be 0 on a boundary with strictly in-going/out-going field.

\begin{equation}
    (\mB_{av,0pad})_{i,:} = \frac{1}{2} \left(
    \frac{|\mF^+_{i,:}|}{||\mF^+_{i,:}||_{L^1} + \epsilon} +
    \frac{|\mF^-_{i,:}|}{||\mF^-_{i,:}||_{L^1} + \epsilon}
    \right)
\end{equation}

Following the same argument, we define $\mB_{dx,0pad}$ below, where either the forward or backward term is ignored. The diagonal term is also removed at the boundary so that the result is a center derivative equal to the subtraction of the forward term with the 0-term on the back (or vice-versa), instead of a forward derivative.

\begin{equation}
\begin{aligned}
\mB_{dx-0pad}(\mF)_{i,:} &= 
\begin{cases}
    \mF_{i,:}'^+ & \text{if } \sum_j{\mF_{i,j}'^-} = 0 \\
    \mF_{i,:}'^-  & \text{if } \sum_j{\mF_{i,j}'^+} = 0 \\
  \frac{1}{2}\left(\mF_{i,:}'^+ + \mF_{i,:}'^- - \text{diag}\left(\sum_{j}{\mF'^+_{:,j} + \mF'^-_{:,j}} \right )_{i,:}\right), & \text{otherwise}
\end{cases}
\\
\mF_{i,:}'^+ &= 
\frac{\mF_{i,:}^+}{||\mF_{i,:}^+||_{L^1} + \epsilon}
\quad \quad
\mF_{i,:}'^- =
\frac{\mF_{i,:}^-}{||\mF_{i,:}^-||_{L^1} + \epsilon}
\end{aligned}
\end{equation}

\subsection{Extending the radius of the aggregation kernel}
\label{app:radius-r-kernel}

We aim at providing a general radius-$R$ kernel $\mB_R$ that assigns different weights to different subsets of nodes $n_u$ at a distance $R$ from the center node $n_v$.

First, we decompose the matrix $\mB(\mF)$ into positive and negative parts $\mB^\pm(\mF)$ representing the forward and backward steps aggregation in the field $\mF$. 
\begin{equation}\label{eq:B_dx decomposition}
\mB(\mF) = \mB^+(\mF) - \mB^-(\mF)
\end{equation}

Thus, defining $\mB_{fb}^\pm(\mF)_{i,:} = \frac{\mF_{i,:}^\pm}{||\mF_{i,:}||_{L^p}}$, we can find different aggregation matrices by using different combinations of walks of radius $R$. First demonstrated for a grid in theorem \ref{th:general_grid_radius2}, we generalize it in \eqref{eq:radius-kernel} for any graph $G$.

\begin{definition}[General radius $R$ n-directional kernel]
\label{def:radius}
Let $S_n$ be the group of permutations over $n$ elements with a set of directional fields $\mF_i$.
\begin{equation}
\label{eq:radius-kernel}
\mB_R := 
\underbrace{
\sum\limits_{\substack{ 
V = \{v_1, v_2, ..., v_n\}  \in \mathbb{N}^n \\ 
|| V ||_{L^1} \le R , \quad
-R \le v_i \le R \\
}}
}_{
\substack{
\text{Any choice of walk $V$ with at most $R$ steps} \\
\text{using all combinations of $v_1, v_2, ... , v_n$}}
}
\underbrace{
\sum_{\sigma \in S_n}}_{\substack{
\text{optional} \\ 
\text{permutations}}}
a_{V}
\underbrace{
\prod_{j=1}^{N}{
(\mB_{fb}^{sgn(v_{\sigma(j)})}( \mF_{\sigma(j)}))^{|v_{\sigma(j)}|}}
}_{
\text{Aggregator following the steps $V$, permuted by $S_n$}}
\end{equation}
\end{definition}

In this equation, $n$ is the number of directional fields and $R$ is the desired radius. $V$ represents all the choices of walk $\{v_1, v_2, ..., v_n\}$ in the direction of the fields $\{\mF_1, \mF_2, ..., \mF_n \}$. For example, $V=\{3, 1, 0, -2\}$ has a radius $R=6$, with 3 steps \textit{forward} of $\mF_1$, 1 step \textit{forward} of $\mF_2$, and 2 steps \textit{backward} of $\mF_4$. The sign of each $\mB_{fb}^\pm$ is dependant to the sign of $v_{\sigma(j)}$, and the power $|v_{\sigma(j)}|$ is the number of aggregation steps in the directional field $\mF_{\sigma(j)}$. The full equation is thus the combination of all possible choices of paths across the set of fields $\mF_i$, with all possible permutations. Note that we are restricting the sum to $v_i$ having only a possible sign; although matrices don't commute, we avoid choosing different signs since it will likely self-intersect a lower radius walk. The permutations $\sigma$ are required since, for example, the path \textit{up $\rightarrow$ left} is different (in a general graph) than the path \textit{left $\rightarrow$ up}.

This matrix $\mB_R$ has a total of $\sum_{r=0}^R (2n)^r = \frac{(2n)^{R + 1} - 1}{2n - 1}$ parameters, with a high redundancy since some permutations might be very similar, e.g. for a grid graph we have that \textit{up $\rightarrow$ left} is identical to \textit{left $\rightarrow$ up}. Hence, we can replace the permutation $S_n$ by a reverse ordering, meaning that $\prod_j^N{\mB_j} = \mB_N ... \mB_2 \mB_1$. Doing so does not perfectly generalize the radius-$R$ kernel for all graphs, but it generalizes it on a grid and significantly reduces the number of parameters to $\sum_{r=0}^R \sum_{l=1}^{min(n, r)} 2^r \binom{n}{l} \binom{r - 1}{l - 1}$.

\subsection{Arcsine of the eigenvectors}
Since the eigenvectors $\eigvec_i$ are equivalent to the Fourier basis and represent the waves in the graphs, then it is expected that they behave similarity to sine/cosine waves when the graph is similar to a grid. This is further highlighted by the proof that the eigenvectors of a grid are all sines/cosines in appendix \ref{app:proof:cosine_eigvec}.

Hence, when we define the field $\mF$ as $\mF^i = \nabla \eigvec_i$, we must realize that the gradient will be lower near the minima/maxima of the eigenvector, as it is the case with sine/cosine waves. In the paper, we cope with this problem by dividing by the norm of the field $\| \mF \|_{L^1}$ in equations \ref{eq:Bav_simple} and \ref{eq:Bdx_simple}.

Another solution is to use the arcsine of the eigenvectors so that the function eigenvectors become similar to triangle functions and the gradient is almost uniform. However, since the arcsine function works only in the range $[-1,1]$, then we must first normalize the eigenvector by it's maximum, as given by equation \ref{eq:F_asin}.

\begin{equation}
\label{eq:F_asin}
    \mF_\text{asin}^i = \nabla \arcsin{\left( \frac{\eigvec_i}{\max(|\eigvec_i|)}  \right)}
\end{equation}

\section{Appendix - Implementation details}

\subsection{Benchmarks and datasets}
\label{app:benchmarks}
We use a variety of benchmarks proposed by \cite{dwivedi2020benchmarking} and \cite{hu2020open} to test the empirical performance of our proposed methods. In particular, to have a wide variety of graphs and tasks we chose: 
\begin{enumerate}
    \item ZINC, a graph regression dataset from molecular chemistry. The task is to predict a score that is a subtraction of computed properties $logP-SA$, with $logP$ being the computed octanol-water partition coefficient, and $SA$ being the synthetic accessibility score \cite{jin_junction_2018}.
    \item CIFAR10, a graph classification dataset from computer vision \cite{krizhevsky_CIFAR10}. The task is to classify the images into 10 different classes, with a total of 5000 training image per class and 1000 test image per class. Each image has $32\times32$ pixels, but the pixels have been clustered into a graph of $\sim100$ super-pixels. Each super-pixel becomes a node in an \textit{almost} grid-shaped graph, with 8 edges per node. The clustering uses the code from \cite{knyazev2019understanding}, and results in a different number of super-pixels per graph. 
    \item PATTERN, a node classification synthetic benchmark generated with Stochastic Block Models, which are widely used to model communities in social networks. The task is to classify the nodes into 2 communities and it tests the fundamental ability of recognizing specific predetermined subgraphs.
    \item MolHIV, a graph classification benchmark from molecular chemistry. The task is to predict whether a molecule inhibits HIV virus replication or not. The molecules in the training, validation and test sets are divided using a scaffold splitting procedure that splits the molecules based on their two-dimensional structural frameworks.
    \item MolPCBA, a graph classification benchmark from molecular chemistry. It consists of measured biological activities of small molecules generated by high-throughput screening. The dataset consists of a total of 437,929 molecules divided using a scaffold slitting procedure and a set of 128 properties to predict for each.
\end{enumerate}

For the results in figure \ref{fig:results_table}, our goal is to provide a fair comparison to demonstrate the capacity of our proposed aggregators. Therefore, we compare the various methods on both types of architectures using the same hyperparameters tuned in previous works \cite{corso2020principal} for similar networks. The models vary exclusively in the aggregation method and the width of the architectures to keep a set parameter budget. Following the indication of the benchmarks' authors, we averaged the performances of the models on 4 runs with different initialization seeds for the benchmarks from \cite{dwivedi2020benchmarking} (ZINC, PATTERN and CIFAR10) and 10 runs for the ones from \cite{hu2020open} (MolHIV and MolPCBA\footnote{For MolPCBA, due to the computational cost of running models in the large dataset and the relatively low variance, we only used 1 run for the results in figure \ref{fig:results_table}, but 10 runs in those for figure \ref{fig:resultscomparison}}).

For the results in figure \ref{fig:resultscomparison}, we took the fine tuned results of other models from the corresponding public leaderboards by \cite{dwivedi2020benchmarking} and \cite{hu2020open}. For the DGN results we fine tuned the model taking the lowest validation loss across runs with the following hyperparameters (you can also find the fine tuned commands in the documentation of the \href{https://github.com/Saro00/DGN}{code repository}):

\begin{enumerate}
    \item ZINC: weight decay $\in \{ 1\cdot 10^{-5}, 10^{-6}, 3 \cdot 10^{-7}\}$, aggregators $\in \{ (mean, avg_1)$, $(mean, dx_1)$, $(mean, av_1, dx_1)$, $(mean, min, max, av_1)$, $(mean, min, max, dx_1)  \}$ 

    \item CIFAR10: weight decay $\in \{ 3 \cdot 10^{-6} \}$, dropout $\in \{ 0.1, 0.3 \}$, aggregators $\in \{ (mean, av_1, av_2)$, $(mean, dx_1, dx_2)$, $(mean, dx_1, dx_2, av_1, av_2)$, $(mean, max, min, dx_1, dx_2)$, $(mean, max, min, av_1, av_2)\}$

    \item PATTERN: weight decay $\in \{ 0, 10^{-8}\}$, architecture $\in \{ simple, complex\}$, aggregators $\in \{ (mean, av_1)$, $(mean, dx_1)$, $(mean, av_1, dx_1) \}$ 

    \item MolHIV: aggregators $\in \{(mean, dx_1)$, $(mean, av_1)$, $(mean, dx_1, av_1)$, $(mean, max, dx_1)$, $(mean, max, dx_1, av_1)$, $(mean, max, min, av_1, dx_1)\}$, dropout $\in \{ 0.1, 0.3, 0.5 \}$, L $\in \{ 4, 6 \}$
    
    \item for MolPCBA, given we did not start from any previously tuned architecture, we performed a line search with the following hyperparameters:  mix of aggregators $\in \{ mean, max, min, sum, dx_1, dx_2, av_1, av_2 \}$, dropout $\in \{ 0.1, 0.2, 0.3, 0.4 \}$, L $\in \{ 4, 6, 8\}$, weight decay $\in \{ 10^{-7}, 10^{-6}, 3\cdot 10^{-6}, 10^{-5}, 3\cdot 10^{-5} \}$, batch size $\in \{ 128. 512. 2048, 3072 \}$, learning rate $\in \{ 10^{-2}, 10^{-3}, 5\cdot 10^{-4}, 2\cdot 10^{-4} \}$, learning rate patience $\in \{ 4, 6, 8 \}$, learning rate reduce factor $\in \{ 0.5, 0.8 \}$, architecture type $\in \{  simple, complex, towers \}$, edge features dimension $\in \{ 0, 8, 16, 32 \}$

\end{enumerate}

In CIFAR10 it is impossible to numerically compute a deterministic vector field with eigenvectors due to the multiplicity of $\lambda_1$ being greater than 1. This is caused by the symmetry of the square image, and is extremely rare in real-world graphs. Therefore, we used as underlying vector field the gradient of the coordinates of the image. Note that these directions are provided in the nodes' features in the dataset and available to all models, that they are co-linear to the eigenvectors of the grid as per lemma \ref{lemma:cosine_eigvec}, and that they mimic the inductive bias in CNNs.

\subsection{Implementation and computational complexity}
\label{app:computation-complexity}
Unlike several more expressive graph networks \cite{kondor2018covariant, maron2018invariant}, our method does not require a computational complexity superlinear with the size of the graph. The calculation of the first $k$ eigenvectors during pretraining, done using Lanczos method \cite{lanczos1950iteration} and the sparse module of Scipy, has a time complexity of $O(Ek)$ where $E$ is the number of edges. During training the complexity is equivalent to a $m$-aggregator GNN $O(Em)$ \cite{corso2020principal} for the aggregation and $O(Nm)$ for the MLP.

To all the architectures we added residual connections \cite{he2016deep}, batch normalization \cite{ioffe2015batch} and graph size normalization \cite{dwivedi2020benchmarking}.

For some of the datasets with non-regular graphs, we combine the various aggregators with logarithmic degree-scalers as in \cite{corso2020principal}.

An important thing to note is that, for dynamic graphs, the eigenvectors need to be re-computed dynamically with the changing edges. Fortunately, there are random walk based algorithms that can estimate $\eigvec_1$ quickly, especially for small changes to the graph \cite{doshi_fiedler_2020}. In the current empirical results, we do not work with dynamic graphs.

To evaluate the difficulty of computing the eigenvectors on very large graphs, we decided to load the COLLAB dataset comprising of a single graph with 235k nodes and 2.35M edges \cite{dwivedi2020benchmarking}. Computing it's first 6 eigenvectors using the scipy \textit{eigsh} function with machine precision took 25.5 minutes on an Intel\textregistered \space Xeon\textregistered \space CPU @ 2.20GHz. This is acceptable, knowing that a general training time can take hours, and that the result can be cached and reused during debugging and hyper-parameter optimization.

\subsection{Running time} \label{app:running_time}

The precomputation of the first four eigenvectors for all the graphs in the datasets takes $38s$ for ZINC, $96s$ for PATTERN and $120s$ for MolHIV on CPU. Table \ref{tab:running_time} shows the average running time on GPU for all the various model from figure \ref{fig:results_table}. On average, the epoch running time is 15\% slower for the DGN compared to the mean aggregation, but a faster convergence for DGN means that the total training time is on average 2\% faster for DGN.

\begin{table}[h]
\caption{Average running time for the non-fine tuned models from figure \ref{fig:results_table}. Each entry represents average time per epoch / average total training time. For the first four datasets, each of the models has a parameter budget $\sim100k$ and was run on a Tesla T4 (15GB GPU). The \textit{avg increase} row is the average of the relative running time of all rows compared to the \textit{mean} row, with a negative value meaning a faster running time.  }
\label{tab:running_time}
\begin{center}
\begin{tabular}{c|ccc|cc}
\hline
                     & \multicolumn{3}{c}{\textbf{ZINC}}                       & \multicolumn{2}{|c}{\textbf{PATTERN}} \\ \hline
\textbf{Aggregators} & \textbf{Simple} & \textbf{Complex} & \textbf{Complex-E} & \textbf{Simple}  & \textbf{Complex}  \\ \hline
mean                 & 3.29s/1505s     & 3.58s/1584s      & 3.56s/1654s        & 153.1s/10154s    & 117.8s/9031s      \\
mean dx$_1$             & 3.86s/1122s     & 3.77s/1278s      & 4.22s/1371s        & 144.9s/8109s     & 127.2s/8417s      \\
mean dx$_1$ dx$_2$         & 4.23s/1360s     & 4.55s/1560s      & 4.63s/1680s        & 153.3s/8057s     & 167.9s/9326s      \\
mean av$_1$             & 3.68s/1297s     & 3.84s/1398s      & 3.92s/1272s        & 128.0s/8680s     & 88.1s/7456s       \\
mean av$_1$ av$_2$         & 3.95s/1432s     & 4.03s/1596s      & 4.07s/1721s        & 134.2s/8115s     & 170.4s/11114s     \\
mean dx$_1$ av$_1$         & 3.89s/1079s     & 4.09s/1242s      & 4.58s/1510s        & 118.6s/6221s     & 144.2s/9112s      \\
\hline
avg increase         & +19\%/-16\%     & +13\%/-11\%      & +20\%/-9\%        & -11\%/-23\%     & +18\%/+1\%      \\
\hline
\end{tabular}

\vspace{10pt}

\begin{tabular}{c|cc|c|cc}
\hline
                     & \multicolumn{2}{c|}{\textbf{CIFAR10}} & \textbf{MolHIV} & \multicolumn{2}{c}{\textbf{MolPCBA}}\\ \hline
\textbf{Aggregators} & \textbf{Simple}  & \textbf{Complex}  & \textbf{Simple} & \textbf{Complex} & \textbf{Complex-E} \\ \hline
mean                 & 83.6s/10526s     & 78.7s/10900s      & 11.4s/2189s     & 279s/30128s & 356s/38126s \\
mean dx$_1$             &                  &                   & 12.6s/2348s    & 304s/34129s & 461s/43419s \\
mean dx$_1$ dx$_2$         & 98.4s/8405s      & 100.9s/5191s      & 14.1s/2345s    & 314s/36581s & 334s/38363s \\
mean av$_1$             &                  &                   & 12.2s/2177s    & 297s/30316s & 436s/54545s  \\
mean av$_1$ av$_2$         & 117.1s/12834s    & 89.5s/14481s      & 13.9s/2150s    & 315s/42297s & 333s/36641s \\
mean dx$_1$ av$_1$         &                  &                   & 14.0s/2070s    & 326s/37523s & 461s/59109s \\ 
\hline
avg increase         & +29\%/+1\%     & +21\%/-10\%      & +17\%/+1\%  & +12\%/+20\%  &  +14\%/+22\%\\
\hline
\end{tabular}

\end{center}
\end{table}

\subsection{Eigenvector multiplicity} \label{app:eig_mult}

The possibility to define equivariant directions using the low-frequency Laplacian eigenvectors is subject to the uniqueness of those vectors.
When the dimension of the eigenspaces associated with the lowest eigenvalues is $1$, the eigenvectors are defined up to a constant factor. In section \ref{sec:grad_eig}, we propose the use of unit vector normalization and an absolute value to eliminate the scale and sign ambiguity. When the dimension of those eigenspaces is greater than $1$, it is not possible to define equivariant directions using the eigenvectors. 

Fortunately, it is very rare for the Laplacian matrix to have repeated eigenvalues in real-world datasets.
We validate this claim by looking at ZINC and PATTERN datasets where we found no graphs with repeated Fiedler vector and only one graph out of 26k with multiplicity of the second eigenvector greater than 1.

When facing a graph that presents repeated Laplacian eigenvalues, we propose to randomly shuffle, during training time, different eigenvectors randomly sampled in the eigenspace. This technique will act as a data augmentation of the graph during training time allowing the network to train with multiple directions at the same time.

\section{Appendix - Mathematical proofs}

\subsection{Proof for theorem \ref{th:dir_smooth} (\nameref{th:dir_smooth})}
\label{app:proof:dir_smooth}
\Paste{th:dir_smooth}

\begin{proof}
This should be a simple proof, that if we want a weighted average of our neighbours, we simply need to multiply the weights by each neighbour, and divide by the sum of the weights. Of course, the weights should be positive.

\end{proof}

\subsection{Proof for theorem \ref{th:dir_dx} (\nameref{th:dir_dx})}
\label{app:proof:dir_dx}
\Paste{th:dir_dx}

\begin{proof}
Since $\mF$ rows have unit $L^1$ norm, $\hat{\mF} = \mF$. The $i$-th coordinate of the vector $\left( \mF - \mathrm{diag} \left(\sum_j \mF_{:,j} \right)\right)\vx$ is 

\begin{align*}
\left(\mF \vx-\mathrm{diag} \left(\sum_j \mF \right)\vx \right)_i &= \sum_j \mF_{i,j}\vx(j) - \left( \sum_{j}\mF_{i,j}  \right)\vx(i) \\
&= \sum_{j:(i,j)\in E}(\vx(j)-\vx(i))\mF_{i,j}\\
&=D_\mF\ \vx(i)
\end{align*}
\end{proof}

\subsection{Proof of theorem \ref{th:reduced-diffusion-distance} (\nameref{th:reduced-diffusion-distance})}
\label{app:proof:reduced-diffusion-distance}

\Paste{th:reduced-diffusion-distance}

Recall that $p_k(x,y) = (D^{-1}A)^k_{x,y}$ is the discrete heat kernel at step $k$, $q_t(x,y) = \sum_{k \geq 0} \frac{e^{-t}t^k}{k!}p_k(x,y)$ is the continuous heat kernel at time $t$.
In \cite{barlow_2017}, it is shown that the continuous heat kernel is computed by 
$q_t(x,y) = e^{-t\mL_\text{norm}}$.
Following \cite{COIFMAN20065} we can diagonalise $q_t$ to get the identity
\begin{equation}\label{eq:diffusion-distance}
d_t(x,y) = \left( \sum_{i=1}^{n-1} e^{-2t\lambda_i} \Big( \eigvec_i(x) - \eigvec_i(y)\Big)^2 \right)^\frac{1}{2}
\end{equation}
The inequality $d_t(x',y) < d_t(x,y)$ is equivalent to
\begin{equation}\label{eq:inegalité-distance-diffusion}
\sum_{i=2}^{n-1}e^{-2t\lambda_i} \left(\Big( \eigvec_i(x') - \eigvec_i(y) \Big)^2 - \Big( \eigvec_i(x) - \eigvec_i(y) \Big)^2 \right) < e^{-2t\lambda_1}\left( \Big( \eigvec_1(x) - \eigvec_1(y) \Big)^2 - \Big( \eigvec_1(x') - \eigvec_1(y) \Big)^2\right)
\end{equation}
The term on the left is bounded above by 
\[\sum_{i=2}^{n-1}e^{-2t\lambda_i} \left|\Big( \eigvec_i(x') - \eigvec_i(y) \Big)^2 - \Big( \eigvec_i(x) - \eigvec_i(y) \Big)^2 \right|
\]
and this last term is in turn bounded above by
\[e^{-2t\lambda_2}\sum_{i=2}^{n-1} \left|\Big( \eigvec_i(x') - \eigvec_i(y) \Big)^2 - \Big( \eigvec_i(x) - \eigvec_i(y) \Big)^2 \right|
\]
Inequality \ref{eq:inegalité-distance-diffusion} will then hold if
\[e^{-2t\lambda_2}\sum_{i=2}^{n-1} \left|\Big( \eigvec_i(x') - \eigvec_i(y) \Big)^2 - \Big( \eigvec_i(x) - \eigvec_i(y) \Big)^2 \right| < e^{-2t\lambda_1}\left( \Big( \eigvec_1(x) - \eigvec_1(y) \Big)^2 - \Big( \eigvec_1(x') - \eigvec_1(y) \Big)^2\right)
\]
and this is equivalent to
\[\frac{1}{2(\lambda_1-\lambda_2)}\log\left( \frac{\left( \Big( \eigvec_1(x) - \eigvec_1(y) \Big)^2 - \Big( \eigvec_1(x') - \eigvec_1(y) \Big)^2\right)}{\sum_{i=2}^{n-1} \left|\Big( \eigvec_i(x') - \eigvec_i(y) \Big)^2 - \Big( \eigvec_i(x) - \eigvec_i(y) \Big)^2 \right|}\right) < t
\]
if we take $t$ to be larger than the term on the left the inequality we get $d_t(x',y) < d_t(x,y)$.
\paragraph{}
The constant $C$ in the statement is the constant on the left side of the inequality. It is also interesting to note that $C$ is expected to be positive since the term $\lambda_1-\lambda_2$ is negative and the argument of the $\log$ will most likely be $<1$.

\subsection{Proof for Lemma \ref{lemma:cosine_eigvec} (\nameref{lemma:cosine_eigvec})}
\label{app:proof:cosine_eigvec}

Consider the lattice graph $\Gamma$ of size $N_1 \times N_2 \times ... \times N_n$, that has vertices $\prod_{i=1,...,n} \{1,...,N_i \}$ and the vertices $(x_i)_{i=1,...,n}$ and $(y_i)_{i=1,...,n}$ are connected by an edge iff $|x_i - y_i| = 1$ for one index $i$ and $0$ for all other indices. Note that there are no diagonal edges in the lattice. The eigenvector of the Laplacian of the grid $L(\Gamma)$ are given by $\eigvec_j$.

\begin{lemma}[Cosine eigenvectors]
\label{lemma:cosine_eigvec}
    \Copy{lemma:cosine_eigvec}{
    The Laplacian of $\Gamma$ has an eigenvalue $2-2\cos\left(\frac{\pi}{N_i}\right)$ with the associated eigenvector $\eigvec_j$ that depends only the variable in the $i$-th dimension and is constant in all others, with $\eigvec_j = \mathbf{1}_{N_1}\otimes \mathbf{1}_{N_2}\otimes ...\otimes \vx_{1,N_i} \otimes ...\otimes \mathbf{1}_{N_n}$,
    and $\vx_{1,N_i}(j) = \cos\left(\frac{\pi j}{n}-\frac{\pi }{2n}\right)$
}
\end{lemma}

\begin{proof}
First, recall the well known result that the path graph on $N$ vertices $P_N$ has eigenvalues 
\[
\lambda_k = 2-2\cos\left(\frac{\pi k}{n}\right)
\]
with associated eigenvector $\vx_k$ with $i$-th coordinate
\[ \vx_k(i) = \cos\left(\frac{\pi ki}{n}+\frac{\pi k}{2n}\right)
\]

The Cartesian product of two graphs $G=(V_G,E_G)$ and $H=(V_H,E_H)$ is defined as $G\times H= (V_{G\times H}, E_{G\times H})$ with $V_{G\times H} = V_G \times V_H$ and $((u_1,u_2), ((v_1,v_2)) \in E_{G\times H}$ iff either $u_1=v_1$ and $(u_2,v_2) \in E_H$ or $(u_1,v_1) \in V_G$ and $u_2=v_2$. It is shown in \cite{fiedler1973algebraic} that if $(\mu_i)_{i=1,...,m}$ and $(\lambda_j)_{j=1,...,n}$ are the eigenvalues of $G$ and $H$ respectively, then the eigenvalues of the Cartesian product graph $G\times H$ are $\mu_i + \lambda_j$ for all possible eigenvalues $\mu_i$ and $\lambda_j$. Also, the eigenvectors associated to the eigenvalue $\mu_i + \lambda_j$ are $u_i \otimes v_j$ with $u_i$ an eigenvector of the Laplacian of $G$ associated to the eigenvalue $\mu_i$ and $v_j$ an eigenvector of the Laplacian of $H$ associated to the eigenvalue $\lambda_j$. 

Finally, noticing that a lattice of shape $N_1\times N_2\times ... \times N_n$ is really the Cartesian product of path graphs of length $N_1$ up to $N_n$, we conclude that there are eigenvalues $2-2\cos\left(\frac{\pi}{N_i}\right)$. Denoting by $\mathbf{1}_{N_j}$ the vector in $\mR^{N_j}$ with only ones as coordinates, then the eigenvector associated to the eigenvalue $2-2\cos\left(\frac{\pi}{N_i}\right)$ is 
\[\mathbf{1}_{N_1}\otimes \mathbf{1}_{N_2}\otimes ...\otimes \vx_{1,N_i} \otimes ...\otimes \mathbf{1}_{N_n}
\]
where $\vx_{1,N_i}$ is the eigenvector of the Laplacian of $P_{N_i}$ associated to its first non-zero eigenvalue. $2-2\cos\left(\frac{\pi}{N_i}\right)$.
\end{proof}

\subsection{Radius 1 convolution kernels in a grid}
In this section we show any radius 1 convolution kernel can be obtained as a linear combination of the $\mB_{dx}(\nabla \eigvec_i)$ and $\mB_{av}(\nabla \eigvec_i)$ matrices for the right choice of Laplacian eigenvectors $\eigvec_i$. First we show this can be done for 1-d convolution kernels.

\begin{theorem}\label{th:1D_convolution_kernel}\Copy{th:1D_convolution_kernel}{
On a path graph, any 1D convolution kernel of size 3 $k$ is a linear combination of the aggregators $\mB_{av}, \mB_{dx}$ and the identity $\mI$.}
\end{theorem}

\begin{proof}
Recall from the previous proof that the first non zero eigenvalue of the path graph $P_N$ has associated eigenvector $\eigvec_1(i) = \cos(\frac{\pi i}{N} - \frac{\pi}{2N})$. Since this is a monotone decreasing function in $i$, the $i$-th row of $\nabla \eigvec_1$ will be 
\[(0,...,0,s_{i-1},0,-s_{i+1},0,...,0)
\]
with $s_{i-1} $ and $s_{i+1} > 0$. We are trying to solve 
\[(a\mB_{av} + b\mB_{dx} + c\mathbf{Id})_{i,:} = (0,...,0,x,y,z,0,...,0)\]
with $x,y,z,$ in positions $i-1, i$ and $i+1$. This simplifies to solving
\[a\frac{1}{\|s\|_{L^1}}|s| + b\frac{1}{\|s\|_{L^2}}s + c(0,1,0) = (x,y,z)
\]
with $s = (s_{i-1},0,-s_{i+1})$, which always has a solution because $s_{i-1}, s_{i+1} > 0$. 
\end{proof}

\begin{theorem}[Generalization radius-1 convolutional kernel in a grid]
\label{th:general_grid_radius1}
    \Copy{th:general_grid_radius1}{
    Let $\Gamma$ be the $n$-dimensional lattice as above and let $\eigvec_j$ be the eigenvectors of the Laplacian of the lattice as in theorem \ref{lemma:cosine_eigvec}. Then any radius 1 kernel $k$ on $\Gamma$ is a linear combination of the aggregators $\mB_{av}(\eigvec_i), \mB_{dx}(\eigvec_i$) and $\mI$.
}
\end{theorem}

\begin{proof}
This is a direct consequence of \ref{th:1D_convolution_kernel} obtained by adding $n$ 1-dimensional kernels, with each kernel being in a different axis of the grid as per Lemma \ref{lemma:cosine_eigvec}. See figure \ref{fig:cnn-gnn-generalization} for a visual example in 2D. 

\end{proof}

\subsection{Proof for theorem \ref{th:general_grid_radius2} (\nameref{th:general_grid_radius2})}
\label{app:proof:general_grid_radius2}
\Paste{th:general_grid_radius2}

\begin{proof}
For clarity, we first do the 2 dimensional case for a radius 2, then extended to the general case. Let $k$ be the radius 2 kernel on a grid represented by the matrix
\[\va_{5\times5}=
\left(
\begin{matrix}
0 & 0 & a_{-2,0} & 0 & 0 \\
0 & a_{-1,-1}& a_{-1,0}& a_{-1,1}& 0 \\
a_{0,-2}& a_{0,-1}& a_{0,0}& a_{0,1}& a_{0,2}\\
0& a_{1,-1}& a_{1,0}& a_{1,1}& 0\\
0& 0& a_{2,0}& 0& 0
\end{matrix}
\right)
\]
since we supposed the $N_1 \times N_2$ grid was such that $N_1>N_2$, by theorem \ref{lemma:cosine_eigvec}, we have that $\eigvec_1$ is depending only in the first variable $x_1$ and is monotone in $x_1$. Recall from \ref{lemma:cosine_eigvec} that 
\[\eigvec_1(i) =  \cos\left(\frac{\pi i}{N_1}+\frac{\pi }{2N_1}\right)
\]
The vector $\frac{N_1}{\pi}\nabla \arccos(\eigvec_1)$ will be denoted by $\mF_1$ in the rest. Notice all entries of $\mF_1$ are $0$ or $\pm1$. Denote by $\mF_2$ the gradient vector $\frac{N_2}{\pi}\nabla \arccos(\eigvec_k)$ where $\eigvec_k$ is the eigenvector given by theorem \ref{lemma:cosine_eigvec} that is depending only in the second variable $x_2$ and is monotone in $x_1$ and recall
\[\eigvec_k(i) = \cos\left(\frac{\pi i}{N_2}+\frac{\pi }{2N_2}\right)
\]
For a matrix $\mB$, let $\mB^\pm$ the positive/negative parts of $\mB$, ie matrices with positive entries such that $\mB = \mB^+ - \mB^-$.
Let $\mB_{r1}$ be a matrix representing the radius 1 kernel with weights
 
\[\va_{3\times 3} = 
\left(
\begin{matrix}
0& a_{-1,0}& 0\\
a_{0,-1}& a_{0,0}& a_{0,1}\\
0& a_{1,0}& 0
\end{matrix}
\right)
\]
The matrix $\mB_{r1}$ can be obtained by theorem \ref{th:general_grid_radius1}. Then the radius 2 kernel $k$ is defined by all the possible combinations of 2 positive/negative steps, plus the initial radius-1 kernel.
\[\mB_{r2} = 
\sum\limits_{\substack{ -2 \leq i,j \leq 2\\ |i|+|j| = 2}}
\underbrace{\left(
a_{i,j} 
(\mF_1^{sgn(i)})^{|i|} (\mF_2^{sgn(j)})^{|j|}
\right)}_{\text{Any combination of 2 steps}}
+ \underbrace{\mB_{r1}}_{\text{all possible single-steps}}
\]
with $sgn$ the sign function $sgn(i)=+$ if $i\geq 0$ and $-$ if $i<0$. The matrix $\mB_{r2}$ then realises the kernel $\va_{5\times 5}$.

We can further extend the above construction to $N$ dimension grids and radius $R$ kernels $k$
\[
{
\underbrace{
\sum\limits_{\substack{ 
V = \{v_1, v_2, ..., v_N\} \in \mathbb{N}^n \\ 
|| V ||_{L^1} \le R \\
-R \le v_i \le R \\
}}}_{\text{Any choice of walk $V$ with at most $R$-steps}}
a_{V}
\underbrace{
\prod_{j=1}^{N}{
(\mF_j^{sgn(v_j)})^{|v_j|}}
}_{\text{Aggregator following the steps defined in $V$}}}
\]

with $\mF_j = \frac{N_j}{\pi} \nabla \arccos \eigvec_j$ ,$\eigvec_j$ the eigenvector with lowest eigenvalue only dependent on the $j$-th variable and given in theorem \ref{lemma:cosine_eigvec} and $\prod$ is the matrix multiplication. $V$ represents all the choices of walk $\{v_1, v_2, ..., v_n\}$ in the direction of the fields $\{\mF_1, \mF_2, ..., \mF_n \}$. For example, $V=\{3, 1, 0, -2\}$ has a radius $R=6$, with 3 steps \textit{forward} of $\mF_1$, 1 step \textit{forward} of $\mF_2$, and 2 steps \textit{backward} of $\mF_4$.

\end{proof}

\subsection{Proof for theorem \ref{th:wl-test} (\nameref{th:wl-test})}
\label{app:proof:wl-test}
\Paste{th:wl-test}

\begin{proof}

We will show that (1) DGNs are at least as powerful as the 1-WL test and (2) there is a pair of graphs which are not distinguishable by the 1-WL test which DGNs can discriminate. 

Since the DGNs include the mean aggregator combined with at least an injective degree-scaler, \cite{corso2020principal} show that the resulting architecture is at least as powerful as the 1-WL test.

\begin{figure}[ht]
\centering
 \includegraphics[width=0.6\textwidth]
 {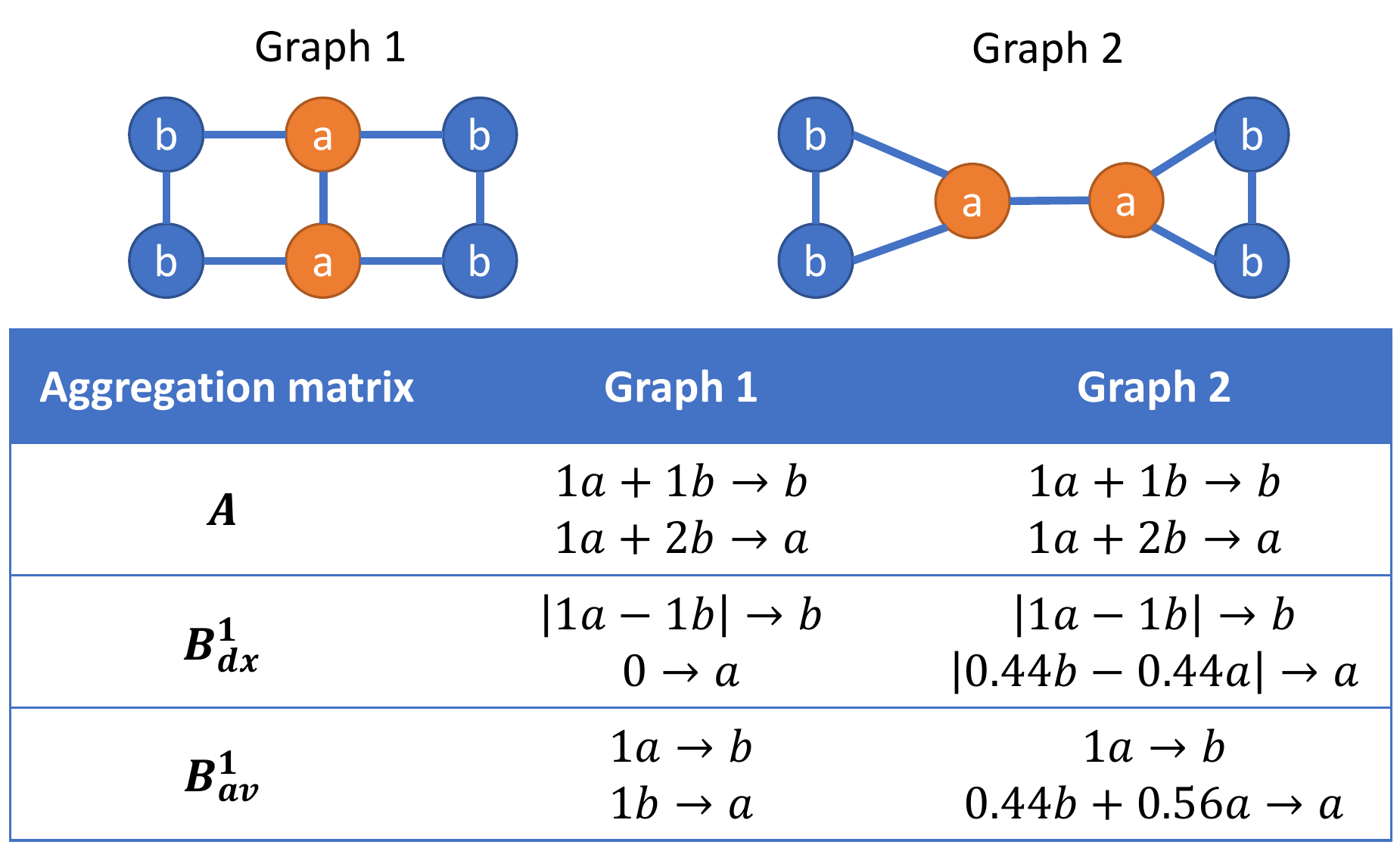}
 \caption{Illustration of an example pair of graphs which the 1-WL test cannot distinguish but DGNs can. The table shows the node feature updates done at every layer. MPNN with mean/sum aggregators and the 1-WL test only use the updates in the first row and therefore cannot distinguish between the nodes in the two graphs. DGNs also use directional aggregators that, with the vector field given by the first eigenvector of the Laplacian matrix, provides different updates to the nodes in the two graphs.}
\label{fig:wl}
\end{figure}

Then, to show that the DGNs are strictly more powerful than the 1-WL test it suffices to provide an example of a pair of graphs that DGNs can differentiate and 1-WL cannot. Such a pair of graphs is illustrated in figure \ref{fig:wl}. 

The 1-WL test (as any MPNN with, for example, sum aggregator) will always have the same features for all the nodes labelled with \textit{a} and for all the nodes labelled with \textit{b} and, therefore, will classify the graphs as isomorphic. DGNs, via the directional smoothing or directional derivative aggregators based on the first eigenvector of the Laplacian matrix, will update the features of the \textit{a} nodes differently in the two graphs (figure \ref{fig:wl} presents also the aggregation functions) and will, therefore, be capable of distinguishing them.

\end{proof}

\end{document}

%% file: math_commands.tex
\usepackage{amsmath,amsfonts,bm}

\def\eqref#1{equation~\ref{#1}}

\def\ceil#1{\lceil #1 \rceil}

\def\1{\bm{1}}

\def\va{{\bm{a}}}

\def\vx{{\bm{x}}}
\def\vy{{\bm{y}}}

\def\mA{{\bm{A}}}
\def\mB{{\bm{B}}}

\def\mD{{\bm{D}}}

\def\mF{{\bm{F}}}
\def\mG{{\bm{G}}}
\def\mH{{\bm{H}}}
\def\mI{{\bm{I}}}

\def\mL{{\bm{L}}}

\def\mR{{\bm{R}}}

\def\mW{{\bm{W}}}

\DeclareMathAlphabet{\mathsfit}{\encodingdefault}{\sfdefault}{m}{sl}
\SetMathAlphabet{\mathsfit}{bold}{\encodingdefault}{\sfdefault}{bx}{n}

\newcommand{\R}{\mathbb{R}}

\DeclareMathOperator{\diver}{\text{div}}